\documentclass{article} 
\usepackage{fontenc, inputenc, calc, indentfirst, fancyhdr, graphicx, epstopdf, ifthen, lineno, float, amsmath, setspace, mathpazo, booktabs, xcolor, microtype, tikz, amsthm, hyperref, url, geometry, newfloat, caption}
\usepackage[square,numbers]{natbib}
\usepackage{subcaption, bm}
\usepackage{xcolor}

\title{Goal-Directed Planning for Habituated Agents by Active Inference Using a Variational Recurrent Neural Network}

\author{Takazumi Matsumoto$^1$ \and Jun Tani$^{2,*}$}

\date{\small $^1$Okinawa Institute of Science and Technology, \texttt{takazumi.matsumoto@oist.jp}\\
$^2$Okinawa Institute of Science and Technology, \texttt{tani1216jp@gmail.com}\\
$^*$Corresponding author}

\begin{document}
\maketitle
%%%%%%%%%%%%%%%%%%%%%%%%%%%%%%%%%%%%%%%%%%
\section*{Abstract}
It is crucial to ask how agents can achieve goals by generating action plans using only partial models of the world acquired through habituated sensory-motor experiences. Although many existing robotics studies use a forward model framework, there are generalization issues with high degrees of freedom. The current study shows that the predictive coding (PC) and active inference (AIF) frameworks, which employ a generative model, can develop better generalization by learning a prior distribution in a low dimensional latent state space representing probabilistic structures extracted from well habituated sensory-motor trajectories. In our proposed model, learning is carried out by inferring optimal latent variables as well as synaptic weights for maximizing the evidence lower bound, while goal-directed planning is accomplished by inferring latent variables for maximizing the estimated lower bound. Our proposed model was evaluated with both simple and complex robotic tasks in simulation, which demonstrated sufficient generalization in learning with limited training data by setting an intermediate value for a regularization coefficient. Furthermore, comparative simulation results show that the proposed model outperforms a conventional forward model in goal-directed planning, due to the learned prior confining the search of motor plans within the range of habituated trajectories.

\emph{Keywords: goal directed planning; active inference; predictive coding; variational Bayes; recurrent neural network}

%%%%%%%%%%%%%%%%%%%%%%%%%%%%%%%%%%%%%%%%%%
\section{Introduction}
It is generally assumed that agents can never access or acquire complete models of the world which they are interacting with \cite{gabaix14, selten90}. This is because the amount of experience accumulated by interacting with the world in a finite time is limited, and usually the world itself is also dynamically changing. Under such conditions, agents with higher cognitive capability, such as humans, seem to be able to generate feasible goal-directed actions by mentally imaging possible behavioral plans using only partially developed models of the world, learning from limited experiences of interaction with the world.

How is this possible? In addressing this problem, the current paper proposes a novel model for goal-directed plan generation referred to as goal-directed latent variable inference (GLean), based on learning by leveraging two related frameworks, \emph{predictive coding} (PC) \cite{rao99, tani99, lee03, friston05, hohwy13, clark15, friston18} and \emph{active inference} (AIF) \cite{friston09, friston10, friston11, buckley17, pezzulo18, oliver19}. In particular, we attempt to show that agents can generate adequate goal-directed behaviors based on learning in the habituated range of the world by conducting simulation studies on the proposed model.

In brain modeling studies, it has long been considered that the brain uses an internal generative model to predict sensory outcomes of its own actions. In this scenario, the fit between the model's prediction and the actual sensation can be improved in two ways. The first is by adapting belief or intention represented by the internal state of the generative model so that the reconstruction error can be minimized \cite{rao99, friston05, hohwy13, clark15}. This corresponds to perception. The second approach is by acting on the environment in a manner such that the resultant sensation can better fit with the model's prediction \cite{friston09, buckley17, pezzulo18}. The former idea has been formulated in terms of PC and the latter by AIF. However, these two frameworks should be considered in unison as perception and action are effectively two sides of the same coin in enactive cognition.

\begin{figure}[!htbp]
\centering
\hspace*{\fill}
\subcaptionbox{\label{fig:fwd_s}}{\includegraphics[width=0.4\textwidth]{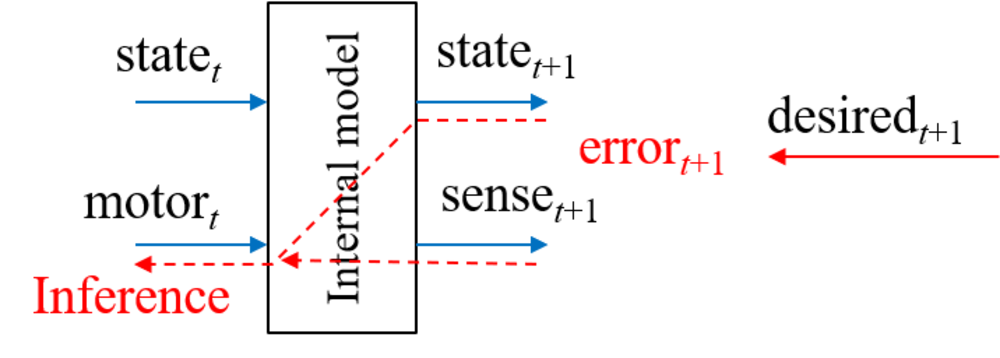}}
\hspace*{\fill}
\subcaptionbox{\label{fig:pcaif_s}}{\includegraphics[width=0.4\textwidth]{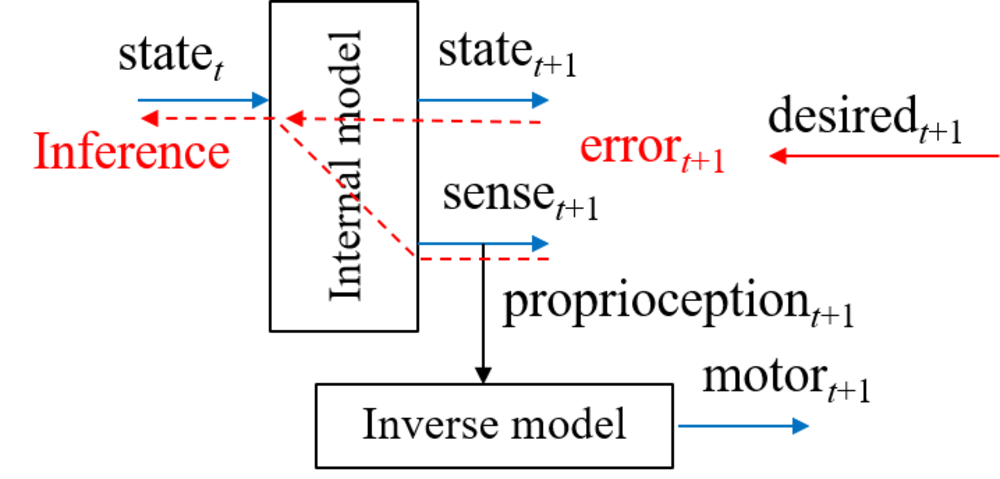}}
\hspace*{\fill}
\vspace{1em}
\caption{\textbf{(a)} The forward model and \textbf{(b)} the predictive coding and active inference framework where $state_{t}$ and $sense_{t+1}$ represent the current latent state and prediction of the next sensory state in terms of the exteroception and proprioception. The predicted proprioception can then be converted into a motor control signal as necessary, such as by using an inverse model as depicted in \textbf{(b)}.}
\end{figure}

Originally, the idea of an internal model for sensory-motor systems was investigated in the study of the forward model (FM) \cite{miall96, kawato90, kawato99} (see Figure \ref{fig:fwd_s}). Although both FM and PC can predict the next latent state and associated sensory inputs, different types of conditioning on the prediction were considered. In FM, the predicted sensory state is conditioned by the current motor commands and the current latent state, while in PC it is conditioned only by the current latent state. In theory, it is possible to infer optimal motor commands for achieving desired states using FM by considering additional cost functions such as jerk minimization, torque minimization, trajectory distance minimization etc. \cite{kawato90}. In practice, however, this inference tends to produce erroneous solutions unless the predictive model learns the outcomes for all possible motor combinations, which is intractable when the motor component has a high degree of freedom.

However, this may not be the case if the PC and AIF frameworks are used in combination as shown in Figure \ref{fig:pcaif_s}. In the PC component, an internal model predicts both the latent state and sensory inputs in terms of the exteroception and proprioception in the next timestep by receiving the current latent state. It can be considered that the sensory sequences are embedded in the latent state space through iterative predictive learning. In the AIF component, an inverse model can be used to map the predicted sensory inputs to motor commands which can realize the predicted sensation. Such an inverse model can be implemented in a straightforward manner by, for example, a PID controller wherein the necessary motor torque to generate expected movements in terms of position and velocity can be computed through error feedback between the predicted proprioception (for e.g., joint angles in a robot) and the outcome.

Given a desired state to be achieved, the optimal motor command at the next timestep can be obtained by first inferring an optimal latent state in the current timestep which can generate the best fit with the desired state with the minimal error. The obtained latent state in the current timestep is mapped to the sensory state of the proprioception and exteroception expected in the next timestep, and the proprioception is finally mapped to a motor command by the inverse model. 

If we assume that agents act on their environment not through all possible combinations of motor command sequences but only a subset of them in terms of habituated trajectories, the effective dimensionality of the sensory space can be reduced drastically. This also results in significant reduction in the required dimensionality of the latent state space which embeds the sensory sequences through learning.

Consequently, the problem of motor planning for achieving a desired state could become tractable when the inference for an optimal latent state can be limited in its relatively low dimensional space. The same, however, cannot be applied in the case of FM as the search for an optimal motor plan cannot be constrained within the range of habituated motor trajectories, as explained previously.

\begin{figure}[!htbp]
\centering
\hspace*{\fill}
\subcaptionbox{\label{fig:fwdmodel}}{\includegraphics[width=0.3\textwidth]{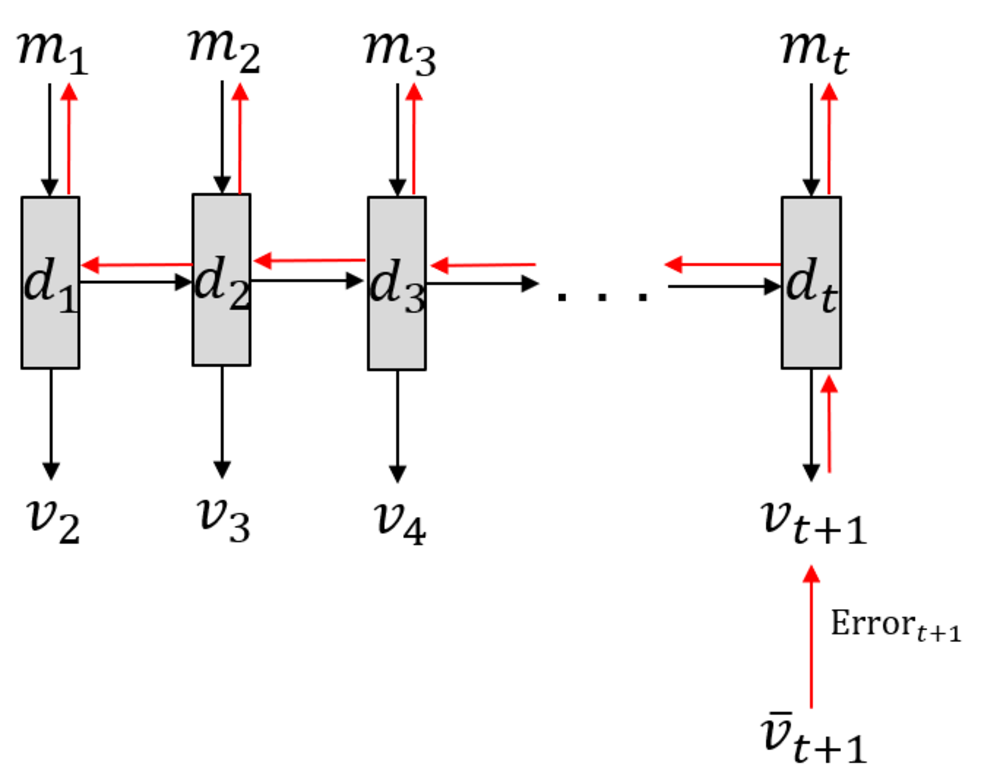}}
\hspace*{\fill}
\subcaptionbox{\label{fig:predcoding0}}{\includegraphics[width=0.345\textwidth]{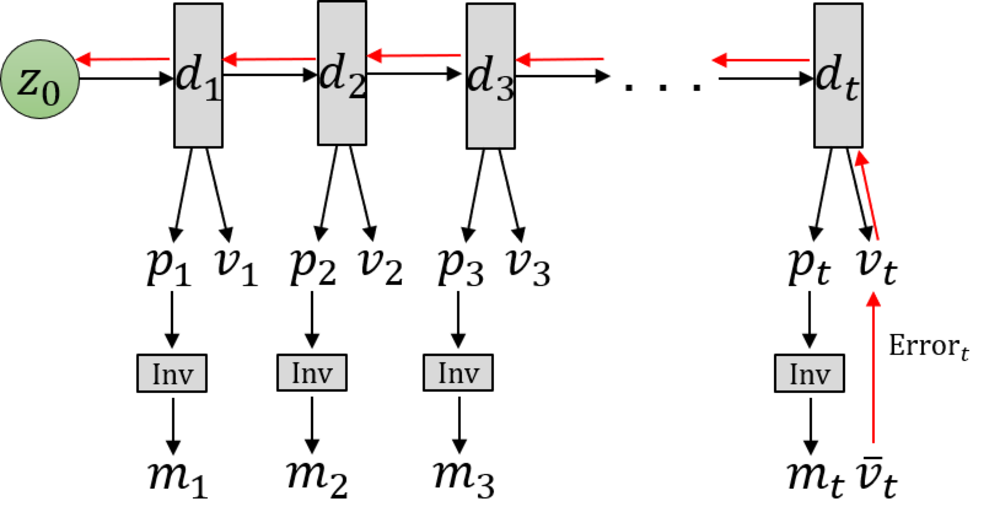}}
\hspace*{\fill}
\subcaptionbox{\label{fig:predcoding1}}{\includegraphics[width=0.3\textwidth]{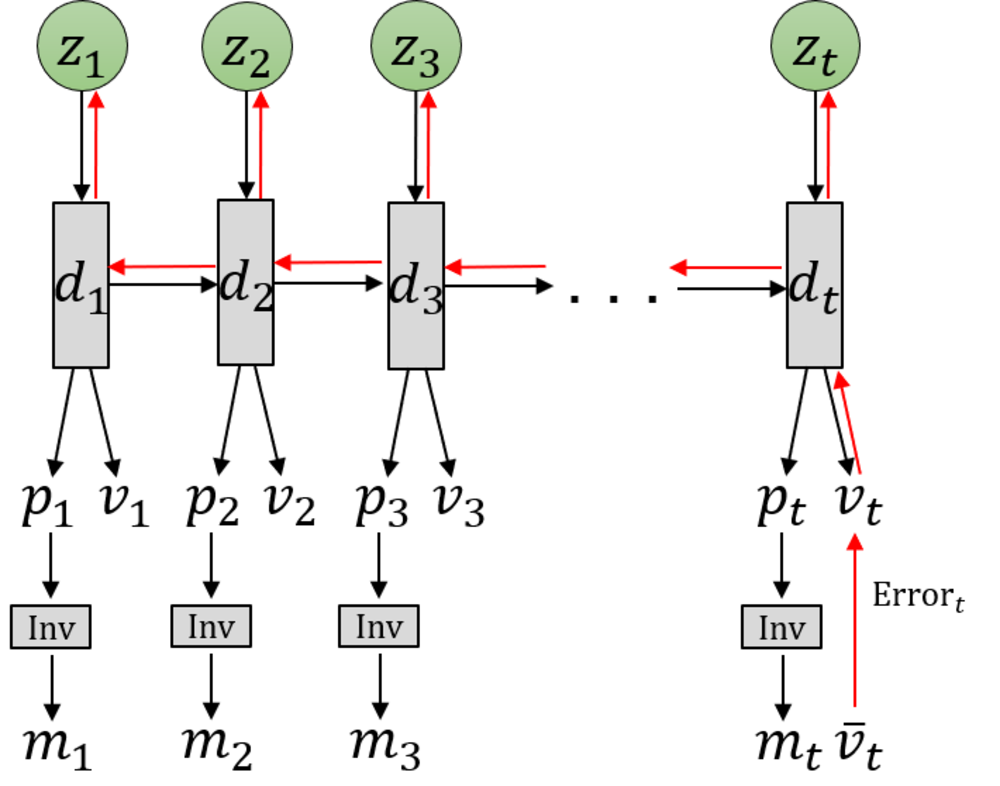}}
\hspace*{\fill}
\vspace{1em}
\caption{Three different models for learning-based goal-directed motor planning. \textbf{(a)} The forward model implemented in an RNN, \textbf{(b)} PC and AIF frameworks implemented in an RNN using initial sensitivity by latent random variables at the initial step, either by the stochastic $\bm{z}_t$ or the deterministic $\bm{d}_t$, and \textbf{(c)} the proposed GLean scheme based on the PC and AIF framework implemented in a variational RNN. In each case, the horizontal axis indicates progression through time (left to right). The black arrows represent computation in the forward pass, while the red arrows represent prediction error being propagated during backpropagation through time (BPTT).\label{fig:modelcompare}}
\end{figure}

The GLean scheme proposed in this paper implements the aforementioned considerations using a variational recurrent neural network (RNN) model and tests it in learning-based robot motor planning tasks. In the following, we briefly review how models of learning-based goal-directed planning have been studied and how such prior studies have been extended to the current work. Tani \cite{tani96} proposed a goal-directed planning scheme for a robot navigation task based on FM (see Figure \ref{fig:fwdmodel}). In this model, the latent state was represented by the activity of the context units in a Jordan-type RNN \cite{jordan86}. During action planning, a sequence of discrete actions $m_1,...,m_t$ such as branching or not branching at each encountered branching point in a maze environment can be inferred by minimizing the error between predicted sensory inputs at distal step $v_{t+1}$ and the sensory inputs associated with the given goal $\overline{v}_{t+1}$.

Arie et al. \cite{arie09} developed a model of learning-based planning analogous to the PC and AIF frameworks (see Figure \ref{fig:predcoding0}). In this model, the initial sensitivity characteristics of deterministic dynamic systems is used wherein diverse sensory-motor sequences experienced are embedded into a distribution of the initial latent state of an RNN model through iterative learning. As such, learning a set of sensory-motor sequences is conducted by means of adapting two different types of variables---connectivity weights of the RNN which are shared by all sequences, and the initial state which is individually adapted for each sequence. After learning with a given initial latent state $\bm{d}_0$, the corresponding sequence consisting of the exteroception $v_{1...t}$ and the proprioception $p_{1...t}$ is generated. By feeding the predicted proprioception at each timestep $p_t$ as the target body posture to the inverse model, the corresponding motor command $m_t$ can be generated. In planning mode, the initial state is inferred such that the distal state in a generated sensory-motor sequence can agree with the desired goal state with minimal error. The inferred initial state represents an intention or belief to generate a motor program reaching the goal state.

Similar work by Choi et al. \cite{choi18} employed a deterministic RNN architecture to accomplish goal-directed motor planning with visual predictions for robotic tasks by searching in this initial state space. In this case, while the network was able demonstrate adequate generalization for simple tasks such as touching a point with a robot arm, the success rate was considerably reduced in a more complex grasp and place task. Recently, Jung et al. \cite{jung19} extended this model by allowing random variables $\bm{z}_0$ with mean and variance to represent the initial states for the purpose of extracting a probabilistic distribution among trained sensory-motor sequences (see Figure \ref{fig:predcoding0}). The experimental results using this model for a task of stacking multiple objects by a robot arm showed that this scheme of using random variables for the initial latent state is beneficial in terms of generalization in both training and motor plan generation.

In this current paper, we propose a further development of the aforementioned model using the framework of PC and AIF to tackle the issue of learning-based goal-directed motor planning by expanding upon the variational Bayes approach. The main purpose of our proposed GLean scheme is to enable the network to learn to extract the transition probability distribution of the latent state at each timestep as a sequence prior \cite{chung15} and to utilize it for generating goal-directed plans with improved generalization. For this purpose, we utilize a recently proposed variational RNN known as the predictive-coding inspired variational RNN (PV-RNN) \cite{ahmadi19} for implementing the PC and AIF frameworks such that the latent state at each timestep is represented by both a deterministic variable $\bm{d}_t$ and a random variable $\bm{z}_t$ as shown in Figure \ref{fig:predcoding1}. Learning of the model is accomplished by maximizing the evidence lower bound, whereas the estimated lower bound is maximized for goal-directed motor plan generation. Both lower bounds are computed as summations of the accuracy term and the complexity term. A formal description of the model is given in Section \ref{sec:model}.

The proposed model also uses ideas considered in development of the so-called Multiple Timescale RNNs (MTRNN) \cite{yamashita08}, which is built on multiple layers of Continuous Time RNNs (CTRNN) \cite{beer95} wherein higher layers have slower timescale dynamics and lower layers have faster dynamics (note that Figure \ref{fig:modelcompare} shows only a single layer for simplicity). It has been shown that MTRNN enhances development of functional hierarchy among layers. It does so by using the timescale difference by which more abstract representations of action plans are developed in the higher layers while a more detailed representation of sensory-motor patterns develop in the lower layers \cite{nishimoto08}.

In Section \ref{sec:experiments} we evaluate GLean by conducting two sets of simulated experiments. Using a minimal task set, the first experiment examines essential characteristics of the proposed model in learning to generate goal-directed plans. In particular, we investigate the effects that regulating the strength of the complexity term in the lower bound has upon learning performance as well as goal-directed motor plan generation. Furthermore, we compare the difference in planning performance between more habituated goal states and less habituated goal states in order to examine the effect of habituation in learning on goal-directed plan generation.

The second simulation experiment uses a more realistic robotic task employing a model of a real robot and compares the performance between three models depicted in Figure \ref{fig:modelcompare}: FM, PC + AIF with initial state sensitivity, and the proposed GLean scheme. These experiments will clarify how GLean can generate feasible goal-directed plans and the resultant actions. It does so by developing regions of habituation in terms of the sequence prior in the latent state space by means of learning from a limited amount of sensory-motor experiences. 

%%%%%
\section{Model} \label{sec:model}
In this section, we will first present an overview of the PV-RNN model followed by a more detailed explanation of training and planning, including formulation of the evidence lower bound and approximate lower bound used in training and planning respectively. We do not attempt to make an exhaustive derivation of PV-RNN in this paper, rather we focus on the salient points and changes compared to the originally proposed model in \cite{ahmadi19}.

\subsection{Overview of PV-RNN}
\begin{figure}[!htbp]
\centering
\includegraphics[width=0.4\textwidth]{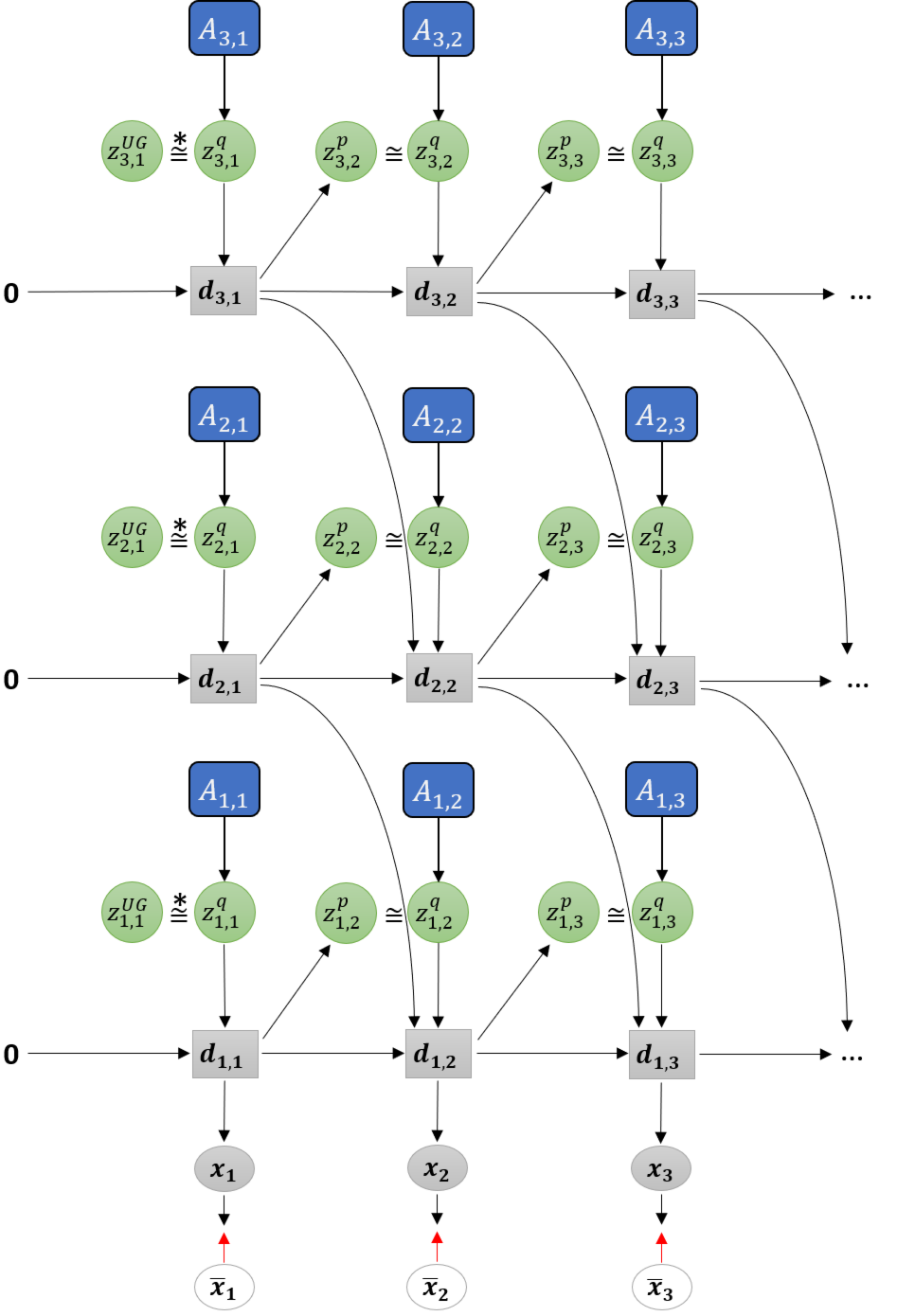}
\caption{Graphical representation of PV-RNN as implemented in this paper}
\label{fig:gp}
\end{figure}

Figure \ref{fig:gp} shows a graphical representation of PV-RNN as implemented in this paper. Note that for generality we denote all the output of the model as $\bm{x}$. Compared to the original PV-RNN, we have made three key changes to the model. The first is at $t=1$, the prior distribution is fixed as a unit Gaussian (depicted as $\bm{z}^{UG}$), which acts as a weighted regularization on initial state sensitivity of the network. This is primarily to improve the stability of learning in certain edge conditions. Note that the deterministic variables are always initialized at zero at $t=0$. In practice, the deterministic variables will tend to learn the mean of the training sequence while the stochastic variables will learn the deviations from the mean.

Secondly, bottom-up connections from lower layers to higher layers have been removed in order to simplify the model. Prediction error from lower layers is still conveyed to higher layers during back-propagation. Additionally, connections between $\bm{z}$ and the output $\bm{x}$ have been removed. Preliminary testing has not shown any degradation of planning performance due to this change.

Finally, connections between $\bm{d}_t$ and the posterior distribution $\bm{z}^q$ have been removed. Thus information from the previous timestep flows between stochastic units only by how close the prior and posterior distributions are, which is regulated by the meta-prior setting. While this change could impact learning performance, it makes inference of the adaptation variables $\bm{A}$ simpler. 

As noted previously, PV-RNN is a variational RNN comprised of deterministic variables $\bm{d}$ and stochastic variables $\bm{z}$. The model infers an approximate posterior distribution $q$ by the prior distribution $p$ by means of error minimization on the generated output $\bm{x}$. The parameterized prior generative model $p_\theta$ is factorized as shown in Equation \ref{eq:genp}.

\begin{equation} \label{eq:genp}
    p_\theta(\bm{x}_{1:T}, \bm{d}_{1:T}, \bm{z}_{1:T} | \bm{d}_0) = \prod_{t=1}^T p_{\theta_x}(\bm{x}_t | \bm{d}_t)  p_{\theta_d}(\bm{d}_t | \bm{d}_{t-1}, \bm{z}_t) p_{\theta_z}(\bm{z}_t | \bm{d}_{t-1})
\end{equation}

Note that unlike in the original implementation of PV-RNN, $\bm{x}$ is not conditioned directly on $\bm{z}$, only through $\bm{d}$, which is a Dirac delta function as defined in Equation \ref{eq:d}.

\begin{equation} \label{eq:d}
    \bm{d}_t =
\begin{cases}
    0 & \text{if } t = 0\\
    f_{\theta_d}(\bm{d}_{t-1}, \bm{z}_t) & \text{if } t > 0
\end{cases}
\end{equation}

$f_{\theta_d}$ is a neural network---MTRNN is used in this paper. $\bm{d}$ is then the output of the MTRNN, which is the internal state $\bm{h}$ after activation. For a multi-layer MTRNN in this model, $\bm{h}$ is calculated as a sum of the value of the stochastic variable $\bm{z}$, the previous timestep output of the current level $l$ and previous timestep output of the next higher level $l+1$ as shown in Equation \ref{eq:cell}.

\begin{equation} \label{eq:cell}
\begin{aligned}
    \bm{d}^l_t &= \text{tanh}(\bm{h}^l_t)\\
    \bm{h}^l_t &= \left(1 - \frac{1}{\tau^l}\right)\bm{h}^l_{t-1} + \frac{1}{\tau^l} \left( \bm{W}^{l,l}_{d,d}\bm{d}^l_{t-1} + \bm{W}^{l,l}_{z,d}\bm{z}^l_t + \bm{W}^{l+1,l}_{d,d}\bm{d}^{l+1}_{t-1} \right)
\end{aligned}
\end{equation}

$\bm{W}$ represent connectivity weight matrices, in this case between layers and between deterministic and stochastic units. Note that at the top layer, $\bm{W}^{l+1,l}_{d,d}\bm{d}^{l+1}_{t-1}$ is omitted.

The prior distribution $p$ of $\bm{z}$ is a Gaussian distribution which depends on $\bm{d}_{t-1}$, except at $t=1$ which does not depend on $\bm{d}_0$ and is fixed as a unit Gaussian. $\bm{\mu}$ and $\bm{\sigma}$ for the prior distribution are obtained from $\bm{d}$ as shown in Equation \ref{eq:p}.

\begin{equation} \label{eq:p}
\begin{aligned}
    p(\bm{z}_1) &= \mathcal{N}(0, I)\\
    p(\bm{z}_t | \bm{d}_{t-1}) &= \mathcal{N}(\bm{\mu}^p_t, (\bm{\sigma}^p_t)^2) \text{ where $t>1$}\\
    \bm{\mu}^p_t &= \text{tanh}(\bm{W}^{l,l}_{d,z,\mu^p}\bm{d}_{t-1})\\
    \bm{\sigma}^p_t &= \exp(\bm{W}^{l,l}_{d,z,\sigma^p}\bm{d}_{t-1})
\end{aligned}
\end{equation}

Based on the reparameterization trick proposed by Kingma and Welling \cite{kingma14}, the latent value $\bm{z}$ for both prior and posterior distributions is a function of $\mu$ and $\sigma$ and a noise sample $\epsilon \sim \mathcal{N}(0, I)$.

\begin{equation}
    \bm{z}_t = \bm{\mu}_t + \bm{\sigma}_t \times \epsilon
\end{equation}

Since computing the true posterior distribution is intractable, the model infers an approximate posterior $q$ of $\bm{z}$ as described in Equation \ref{eq:q}. In PV-RNN, while sensory information $\overline{\bm{x}}$ is not directly available to the network, an adaptation variable $\bm{A}$ is used, so for each training sequence $\overline{\bm{x}}_{1:T}$ there is a corresponding $\bm{A}_{1:T}$. $\bm{A}$ is learned together with the other network parameters during training based on the prediction errors $\bm{e}$ between $\bm{x}$ and $\bm{\overline{x}}$.

\begin{equation} \label{eq:q}
\begin{aligned}
    q(\bm{z}_t | \bm{e}_{t:T}) &= \mathcal{N}(\bm{\mu}^q_t, (\bm{\sigma}^q_t)^2)\\
    \bm{\mu}^q_t &= \text{tanh}(\bm{A}^\mu_t)\\
    \bm{\sigma}^q_t &= \exp(\bm{A}^\sigma_t)
\end{aligned}
\end{equation}

\subsection{Learning with evidence lower bound}
Following from Equation \ref{eq:genp}, we can express the marginal likelihood (evidence) as shown in Equation \ref{eq:genp2}. As the value of $\bm{d}$ is deterministic, if we let $\bm{\tilde{d}_t}$ be the value of $\bm{d_t}$ as described by Equation \ref{eq:d}, then $p_{\theta_d}(\bm{d}_t | \bm{d}_{t-1}, \bm{z}_t)$ is equivalent to a Dirac distribution given by $\delta(\bm{d}_t - \bm{\tilde{d}}_t)$, which allows the integral over $\bm{d}$ to be eliminated.
\begin{equation} \label{eq:genp2}
\begin{aligned}
    p_\theta(\bm{x}_{1:T} | \bm{d}_0) &= \int \int \prod_{t=1}^T p_{\theta_x}(\bm{x}_t | \bm{d}_t) p_{\theta_d}(\bm{d}_t | \bm{d}_{t-1}, \bm{z}_t) p_{\theta_z}(\bm{z}_t | \bm{d}_{t-1}) d\bm{z}d\bm{d}\\
    &= \int \int \prod_{t=1}^T p_{\theta_x}(\bm{x}_t | \bm{\tilde{d}}_t) \delta(\bm{d}_t - \bm{\tilde{d}}_t) p_{\theta_z}(\bm{z}_t | \bm{\tilde{d}}_{t-1}) d\bm{z}d\bm{d}\\
    &= \int \prod_{t=1}^T p_{\theta_x}(\bm{x}_t | \bm{\tilde{d}}_t)  p_{\theta_z}(\bm{z}_t | \bm{\tilde{d}}_{t-1}) d\bm{z}
\end{aligned}
\end{equation}

Factoring the integral, taking the logarithm and refactoring with the parameterized posterior distribution produces an expectation on the posterior distribution as shown in Equation \ref{eq:logevidence}.

\begin{equation} \label{eq:logevidence}
\begin{aligned}
    \log p_\theta(\bm{x}_{1:T} | \bm{d}_0) &= \log \prod_{t=1}^T \int p_{\theta_x}(\bm{x}_t | \bm{\tilde{d}}_t)  p_{\theta_z}(\bm{z}_t | \bm{\tilde{d}}_{t-1}) d\bm{z}_t\\
    &= \sum_{t=1}^T \log \int p_{\theta_x}(\bm{x}_t | \bm{\tilde{d}}_t)  p_{\theta_z}(\bm{z}_t | \bm{\tilde{d}}_{t-1}) d\bm{z}_t\\
    &= \sum_{t=1}^T \log \int p_{\theta_x}(\bm{x}_t | \bm{\tilde{d}}_t) \frac{p_{\theta_z}(\bm{z}_t | \bm{\tilde{d}}_{t-1})}{q_\phi(\bm{z}_t | \bm{e}_{t:T})} q_\phi(\bm{z}_t | \bm{e}_{t:T}) d\bm{z}_t
\end{aligned}
\end{equation}

Finally, by applying Jensen's inequality $\log E[x] \geq E[\log x]$, the variational evidence lower bound (ELBO) $L(\theta, \phi)$ is given in Equation \ref{eq:elbo}.

\begin{equation} \label{eq:elbo}
    L(\theta, \phi) = \sum_{t=1}^T \int \log \Bigg[ p_{\theta_x}(\bm{x}_t | \bm{\tilde{d}}_t) \frac{p_{\theta_z}(\bm{z}_t | \bm{\tilde{d}}_{t-1})}{q_\phi(\bm{z}_t | \bm{e}_{t:T})} \Bigg] q_\phi(\bm{z}_t | \bm{e}_{t:T}) d\bm{z}_t
\end{equation}

Following the concept of free energy minimization \cite{friston06}, ELBO is rewritten in terms of expected log likelihood under the posterior distribution (\emph{accuracy}) and the Kullback-Leibler divergence (KLD) between the posterior and prior distributions (\emph{complexity}) in Equation \ref{eq:femin}. The deterministic value in the expected log likelihood is substituted with all previous stochastic variables by Equation \ref{eq:d} in order to allow optimization of the posterior adaptive values against the training data. For simplicity, we omit the summation over each layer of the RNN and over each training sample.

\begin{equation} \label{eq:femin}
\begin{aligned}
    L(\theta, \phi) &= \sum_{t=1}^T E_{q_\phi(\bm{z}_t | \bm{z}_{1:t-1}, \bm{e}_{t:T})} \big[ p_{\theta_x}(\bm{x}_t | \bm{\tilde{d}}_t) \big] - D_{KL}\big[ q_\phi(\bm{z}_t | \bm{e}_{t:T}) || p_{\theta_z}(\bm{z}_t | \bm{\tilde{d}}_{t-1}) \big]\\
    &= \underbrace{\sum_{t=1}^T E_{q_\phi(\bm{z}_t | \bm{z}_{1:t-1}, \bm{e}_{t:T})} \big[ p_{\theta_x}(\bm{x}_t | \bm{z}_{1:t}) \big]}_\text{Accuracy} - \underbrace{D_{KL}\big[ q_\phi(\bm{z}_t | \bm{e}_{t:T}) || p_{\theta_z}(\bm{z}_t | \bm{z}_{1:t-1}) \big]}_\text{Complexity}
\end{aligned}
\end{equation}

Intuitively, the accuracy term is calculated by the distance between the predicted output $\bm{x}$ and the sensory state or ground truth $\bm{\overline{x}}$. In practice, this is a standard measure such as mean squared error (MSE) or KLD.

In PV-RNN, the meta-prior $w$ is a hyperparameter which affects the degree of regularization (or the tendency to overfit). It is similar to the $\beta$ parameter in VAE \cite{kingma14} although the effect is reversed, that is, in models that assume a prior normal distribution, a larger regularization constant implies a stronger pull toward the normal distribution, reducing complexity and reducing the tendency to overfit. However, as PV-RNN the prior is conditioned on the output of previous timesteps, a larger meta-prior causes the complexity to rise as the output becomes deterministic, resulting in a tendency to overfit training samples. During learning, the meta-prior will affect the approximate posterior distribution and cause it to deviate from the true posterior, while during inference the meta-prior will control how much the prior and approximate posterior will deviate. We explore this effect in the following Section \ref{sec:experiments}.

In this implementation of PV-RNN, the complexity term at $t=1$ is a special case where the prior distribution is a unit Gaussian $\mathcal{N}(0,I)$, and the initial Gaussian weight $w_I$ controls how closely the posterior follows. This has two effects---firstly, the RNN can be made more or less sensitive to the initial state at $t=1$ by adjusting the degree of regularization with a unit Gaussian. Secondly, as it is independent of the meta-prior, it avoids degenerate cases where learning of a probabilistic training set is unsuccessful due to the meta-prior forcing deterministic behavior. From preliminary testing, we found settings of either $w_I = 0.01$ or $w_I = 0.001$ appropriate depending on the data. Additionally, in this implementation of PV-RNN, we use different values of $w$ per layer $l$. For simplicity, summation over timesteps is omitted in Equation \ref{eq:lz}.

\begin{equation} \label{eq:lz}
    \sum^L_{l=1} w^l \cdot D_{KL}\big[ q_\phi(\bm{z}_t | \bm{e}_{t:T}) || p_{\theta_z}(\bm{z}_t | \bm{\tilde{d}}_{t-1}) \big] =
\begin{cases}
    \sum^L_{l=1} w_I \sum_{\bm{\sigma}, \bm{\mu} \in \bm{z}} \log \frac{1}{\bm{\sigma}^{q,l}_t} + \frac{(-\bm{\mu}^{q,l}_t)^2 + (\bm{\sigma}^{q,l}_t)^2}{2} - \frac{1}{2} & \text{if } t = 1\\
    \sum^L_{l=1} w^l \sum_{\bm{\sigma}, \bm{\mu} \in \bm{z}} \log \frac{\bm{\sigma}^{p,l}_t}{\bm{\sigma}^{q,l}_t} + \frac{(\bm{\sigma}^{p,l}_t - \bm{\mu}^{q,l}_t)^2 + (\bm{\sigma}^{q,l}_t)^2}{2(\bm{\sigma}^{p,l}_t)^2} - \frac{1}{2} & \text{if } t > 1
\end{cases}
\end{equation}

In practice, all parameters are optimized by gradient descent, using the Adam optimizer provided by TensorFlow. We note the conditions used in our experiments in Section \ref{sec:experiments}.
% No detail on the update steps are given, since these are handled by automatic differentiation in TF

%%
\subsection{Plan generation with GLean and the estimated lower bound}
\begin{figure}[!htbp]
\centering
\hspace*{\fill}
\subcaptionbox{\label{fig:erpred}}{\includegraphics[width=0.45\textwidth]{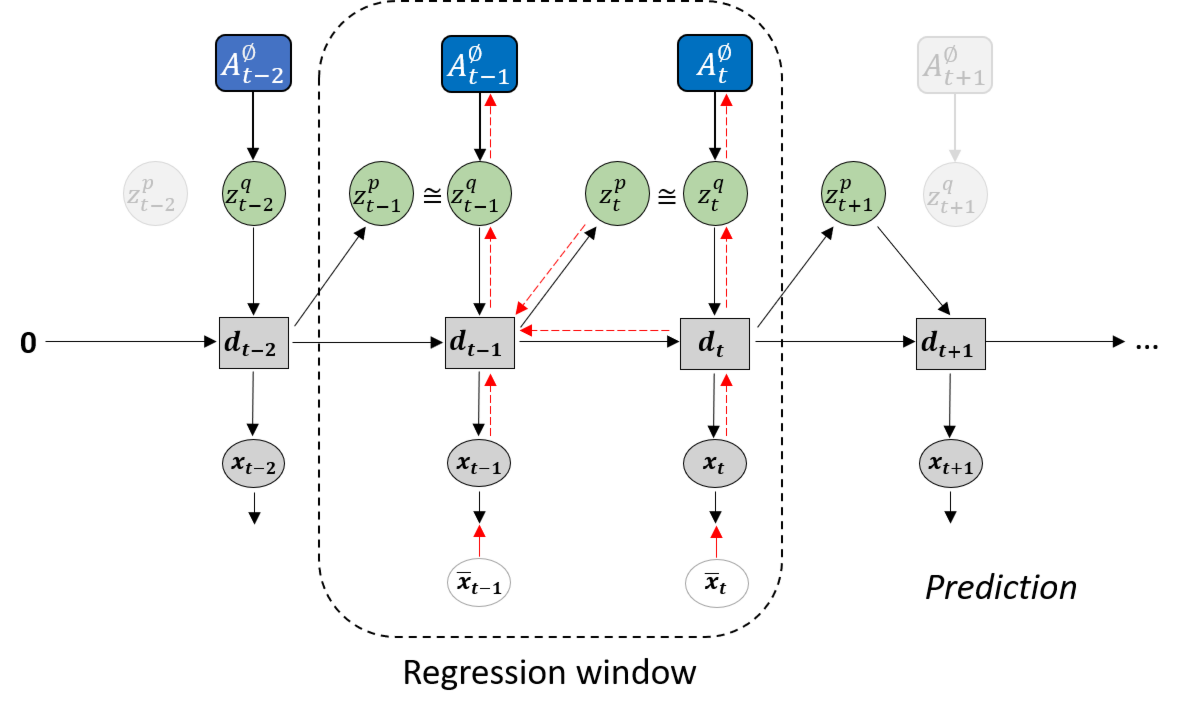}}
\hspace*{\fill}
\subcaptionbox{\label{fig:gdp}}{\includegraphics[width=0.45\textwidth]{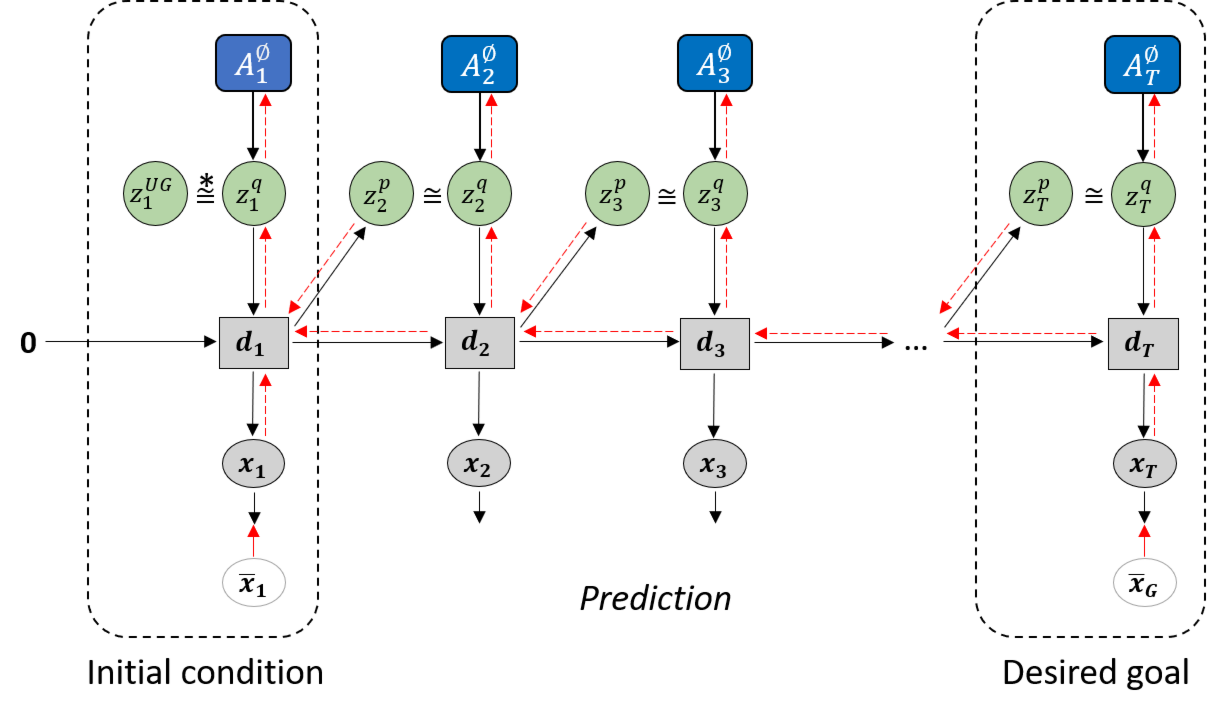}}
\hspace*{\fill}
\vspace{1em}
\caption{Difference in how error regression is employed in \textbf{(a)} future sequence prediction and \textbf{(b)} goal-directed planning. Solid black lines represent the forward generative model while the dashed red lines represent back-propagation through time used to update $A^\emptyset$.}
\end{figure}

Plan generation uses a variation of error regression \cite{tani03} in order to infer the latent variables that minimize the error. However, recent works that utilize error regression \cite{ahmadi19, butz19} employ a regression window in which error is minimized in order to improve future prediction (see Figure \ref{fig:erpred}). GLean attempts to minimize the errors at the initial timestep and the goal timestep (see Figure \ref{fig:gdp}) by maximizing the \emph{estimated lower bound}, shown in Equation \ref{eq:estlbo}.

Compared to ELBO shown in Equation \ref{eq:elbo}, the accuracy term is now calculated as the summation of prediction error in the initial ($t=1$) and distal ($t=T$) steps. In this work, we assume that the distal step is at a fixed point in time; in practice, if the goal is reached early, the agent should remain stationary until the final timestep. The complexity term is also modified such that, except for the first timestep, the posterior distribution is conditioned only on the prediction error at the goal.

\begin{multline} \label{eq:estlbo}
    L_e(\theta, \phi) =  E_{q_\phi(\bm{z}_1 | \bm{e}_1, \bm{e}_T)} \big[ p_{\theta_x}(\bm{x}_1 | \bm{z}_1) \big] + E_{q_\phi(\bm{z}_T | \bm{e}_T)} \big[ p_{\theta_x}(\bm{x}_T | \bm{z}_{1:T}) \big] \\
    - \Big(  D_{KL}\big[ q_\phi(\bm{z}_1 | \bm{e}_1) || p_{\theta_z}(\bm{z}_1) \big] + \sum_{t=2}^T D_{KL}\big[ q_\phi(\bm{z}_t | \bm{e}_T) || p_{\theta_z}(\bm{z}_t | \bm{z}_{1:t-1}) \big] \Big)
\end{multline}

 Note that while the trained model is loaded before plan generation, the adaptive variable $\bm{A}$ is reset to zero (denoted as $\bm{A}^\emptyset$). During plan generation, only the adaptive variable $\bm{A^\emptyset}$ is updated, while all other parameters remain fixed. The implementation of plan generation is largely similar to training, although for practical reasons the number of epochs is significantly reduced. The learning rate, which we refer to as \emph{plan adaptation rate} in the context of plan generation, is raised to compensate for this. In addition, noise sampling is employed by having multiple $\bm{A^\emptyset}$ sequences and selecting for plans with the highest lower bound.

\section{Experiments} \label{sec:experiments}
In order to test GLean, we conducted two experiments with simulated agents. The first experiment was carried out with a virtual mobile agent in a 2D space in order to examine the impact of the meta-prior on learning as well as plan generation outputs. The second experiment used a simulated 8 DOF arm robot carrying out a goal-directed object moving task, and compared GLean to two previously mentioned models---a forward model and a stochastic initial state RNN.

Due to the computational workload of generating long sequences, particularly when executing error regression for plan generation, all plans were generated in an offline manner. This allowed the work to be run in batches on a computer cluster. Similarly, using a simulator to collect data and test the outcomes allowed greater efficiency and automation compared to using real robots. However, in the future, we plan to extend this work to real-time trajectory planning using a physical robot.

As mentioned previously, we implemented PV-RNN and GLean using TensorFlow. The Adam optimizer was used with default parameters, except for learning rate and $\hat{\varepsilon}$ which was set to $1/10$ of learning rate. Additionally we used random dropout of the error signal (i.e. the prediction error $\bm{e}_t$ can either be $\bm{x}-\overline{\bm{x}}$ or $0$).

The source code for GLean is publicly available at \url{https://github.com/oist-cnru/GLean} for both Python 2.7 + TensorFlow 1.x (as tested in this paper) and Python 3.x + TensorFlow 2.x. The tested datasets are also included, together with instructions on how to use the software. 

For these two simulation experiments, we prepared datasets of possible trajectories wherein a portion of the trajectories were used for training of the model and the remaining held back for testing. This provides the ground truth under various conditions including non-goal-directed and goal-directed generation. To evaluate the performance of trajectory generation after the training, we provide both plots of trajectories for qualitative evaluation as well as tables of qualitative measures. For goal-directed plan generation, we judge the quality of the generated outputs by comparing the trajectory to the ground truth trajectory and calculating an average root mean squared error (RMSE) value. The error at the final timestep is also given separately as goal deviation (GD). The average KLD between prior and posterior ($KLD_{pq}$) is stated as an indication of how closely the network is following its trained prior distribution. Note this is equivalent to the complexity term without weighting by the meta-prior.

\subsection{Experiment 1: simulated mobile agent in a 2D space} \label{sec:2dexp}
\begin{figure}[!htbp]
\centering
\hspace*{\fill}
\subcaptionbox{\label{fig:exp1datasetxy}}{\includegraphics[width=0.3\textwidth]{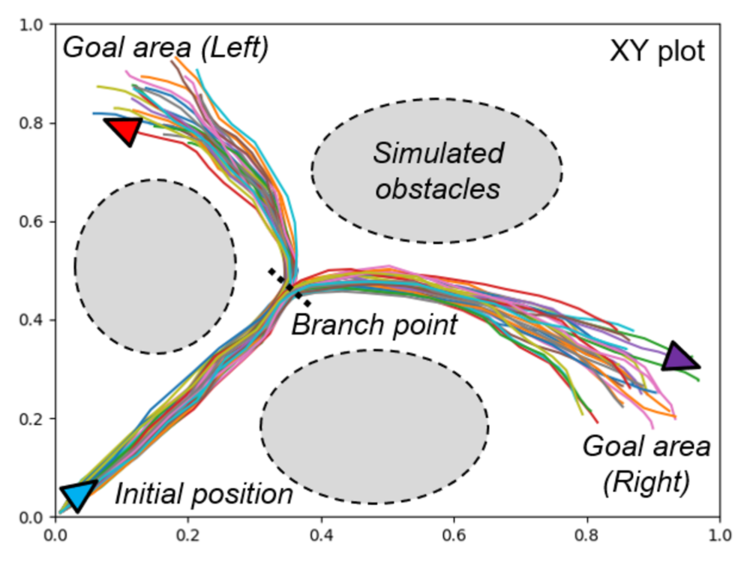}}
\hspace*{\fill}
\subcaptionbox{\label{fig:exp1datasetx}}{\includegraphics[width=0.3\textwidth]{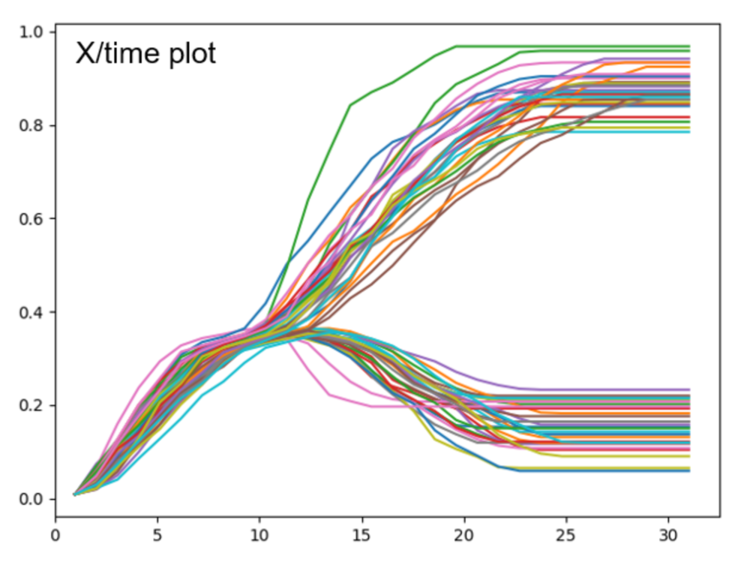}}
\hspace*{\fill}
\subcaptionbox{\label{fig:exp1datasety}}{\includegraphics[width=0.3\textwidth]{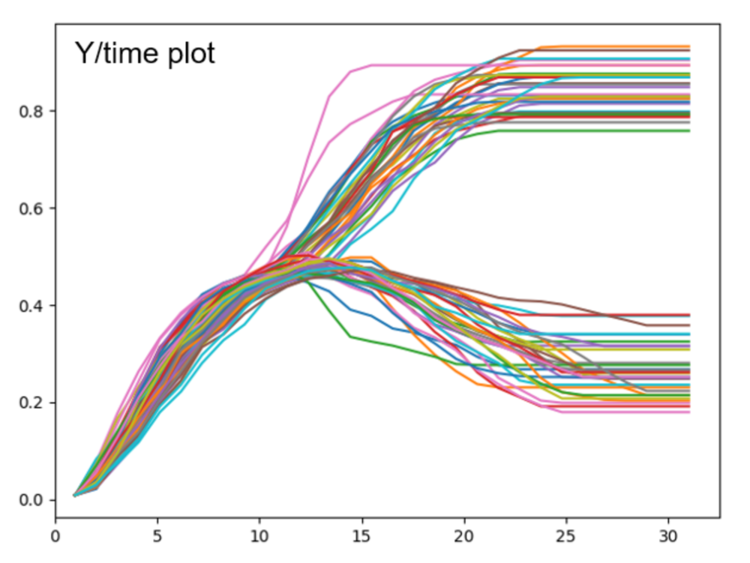}}
\hspace*{\fill}
\vspace{1em}
\caption{Plots of the trajectories prepared for a mobile agent generating goal-directed behaviors in 2D space. \textbf{(a)} XY plot showing the initial position of the agent, the branch point, and the two goal areas, \textbf{(b)} the plot of the X position over time, and \textbf{(c)} the plot of the Y position over time. The branch point is visible at around $t=10$. \label{fig:exp1scenario}}
\end{figure}

To gain a better understanding of the generative capabilities of our model, we first conduct an experiment using a simple simulated agent moving in a 2D space as shown in Figure \ref{fig:exp1scenario}. The agent's position at a given timestep is given by XY coordinates in the range $[0,1]$. The training data consists of hand drawn trajectories resampled to 30 timesteps, starting at $[0,0]$, moving to a central `branch point' at approximately $(0.38, 0.42)$, and then proceeding with a 50/50 chance to one of two goal areas---the top left centered around $(0.2, 0.8)$ and the bottom right centered around $(0.85,0.3)$ without colliding with obstacles shown as grey areas in Figure \ref{fig:exp1datasetxy}, ideally while maintaining a smooth trajectory.

The branch point is reached at approximately $t=10$, with the goal reached between $t=20$ and $t=30$. As the trajectories are hand drawn with a mouse, there is a varying amount of noise in each trajectory and the goal points are also distributed in a fairly uniform distribution. The result is that while the task itself is simple (going from start to one of goal areas), a habituated path of sorts is generated out of the bundle of trajectories drawn.

As noted previously, we use PV-RNN which itself is built on MTRNN. In this experiment, we configure the network to have two layers (note that layer 1 is the bottom layer) with parameters as shown in Table \ref{tbl:exp1rnn}. Neurons refer to the number of deterministic variables, while Z-units refer to the number of stochastic variables. These are kept in a 10:1 ratio as in \cite{ahmadi19}. $\tau$ is the MTRNN time constant, with shorter time constants used in the lower layers which should be more responsive, and longer time constants in the higher layers. The network was trained for 50,000 epochs with a learning rate of 0.001. 

\begin{table}[!htbp]
\caption{PV-RNN parameters for Experiment 1} \label{tbl:exp1rnn}
\centering.
\begin{tabular}{ccc}
\toprule
                          & \multicolumn{2}{c}{MTRNN layer} \\
	                      & \textbf{1}	& \textbf{2} \\
\midrule
Neurons $|\bm{d}^l|$	  & 20			& 10 \\
Z-units	$|\bm{z}^l|$	  & 2			& 1 \\
$\tau$                    & 4           & 8 \\
\bottomrule
\end{tabular}
\end{table}

In order to explore how the meta-prior affects the output in terms of trajectory generation after learning, we prepared three networks trained with different meta-prior values that we have labeled `weak', `intermediate' and `strong' as shown in Table \ref{tbl:exp1w}.

\begin{table}[!htbp]
\caption{Meta-prior settings for the 2D experiment} \label{tbl:exp1w}
\centering
\begin{tabular}{ccc}
\toprule
                                 & \multicolumn{2}{c}{MTRNN layer} \\
\textbf{Meta-prior setting $w$}  & \textbf{1}	& \textbf{2} \\
\midrule
Weak		                     & 0.00001		& 0.000005 \\
Intermediate                     & 0.01 		& 0.005 \\
Strong                           & 0.2          & 0.1 \\
\bottomrule
\end{tabular}
\end{table}

\subsubsection{Prior generation}
\begin{figure}[!htbp]
\centering
\includegraphics[width=0.5\textwidth]{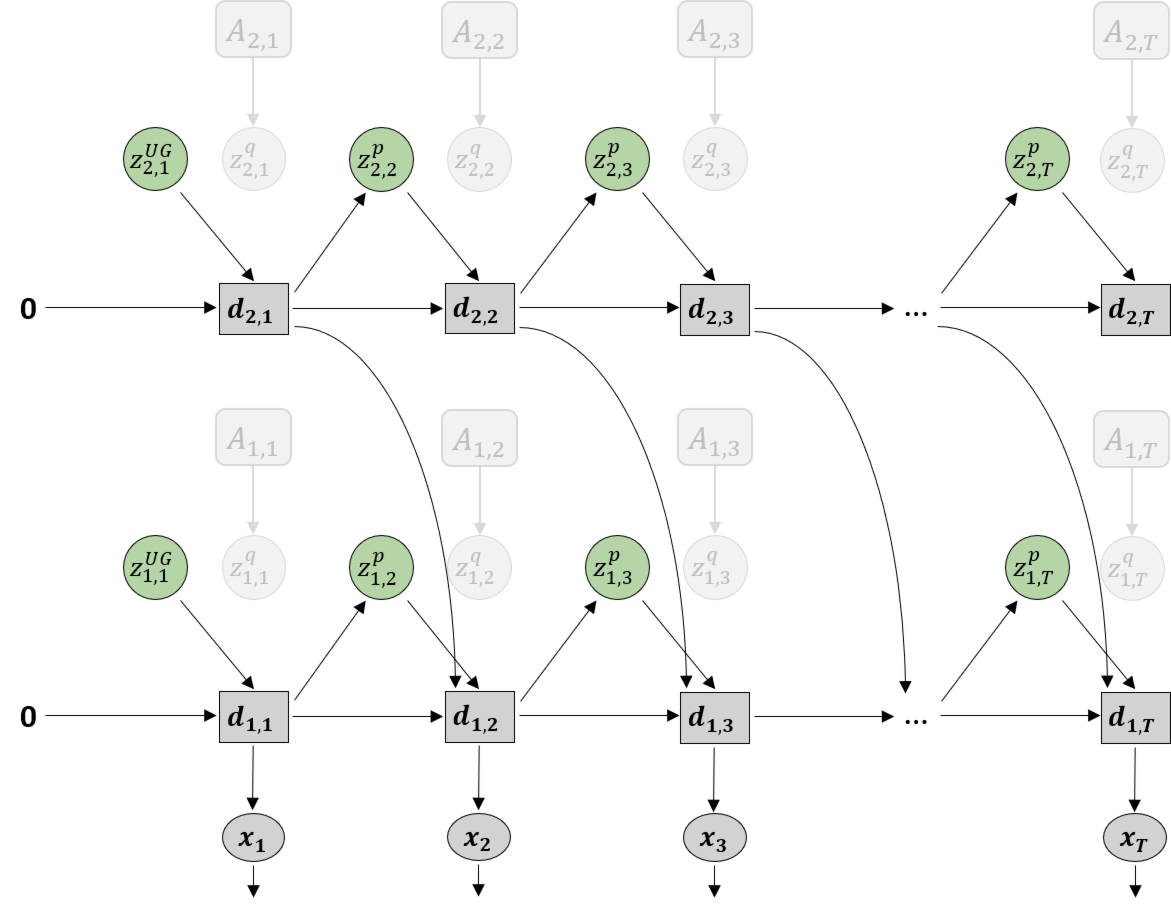}
\caption{Generation using a stochastic initial state (unit Gaussian)}
\label{fig:priorgen}
\end{figure}

To evaluate the ability of the network to learn to extract the probabilistic characteristics latent in the training data, we test prior generation of trajectories using the prior distribution as illustrated in Figure \ref{fig:priorgen}. Since there are no target outputs given and $\bm{z}_1$ is a unit Gaussian and $\bm{d}_0 = 0$, the network is not influenced to go to a particular goal direction. Ideally, the distribution of trajectories generated in this manner should match the training data in terms of distribution between left and right goal areas as the prior generation should represent the distribution of habituated trajectories.

\begin{figure}[!htbp]
\centering
\hspace*{\fill}
\subcaptionbox{\label{fig:undtrain}}{\includegraphics[width=0.3\textwidth]{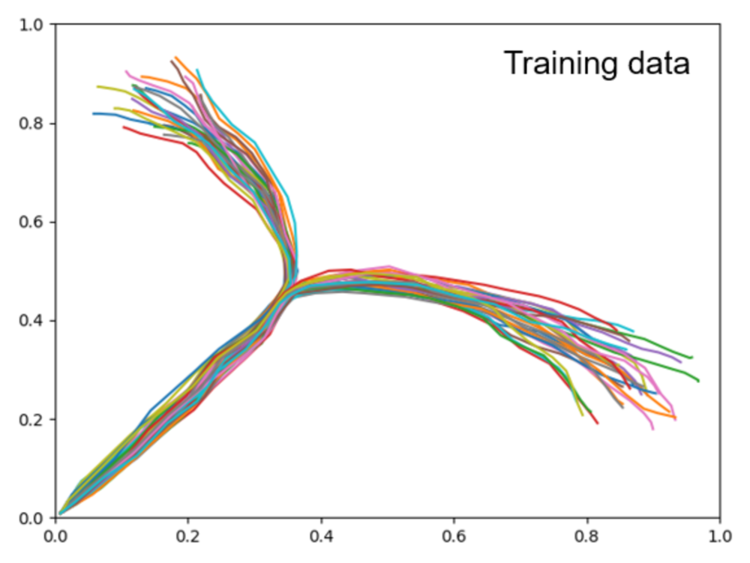}}
\hspace*{\fill}

\vspace{1em}
\hspace*{\fill}
\subcaptionbox{\label{fig:undwtw}}{\includegraphics[width=0.3\textwidth]{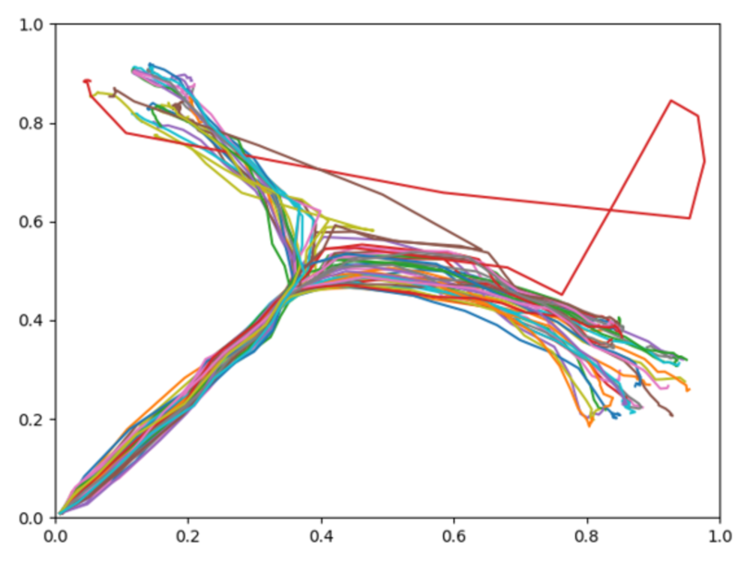}}
\hspace*{\fill}
\subcaptionbox{\label{fig:unditw}}{\includegraphics[width=0.3\textwidth]{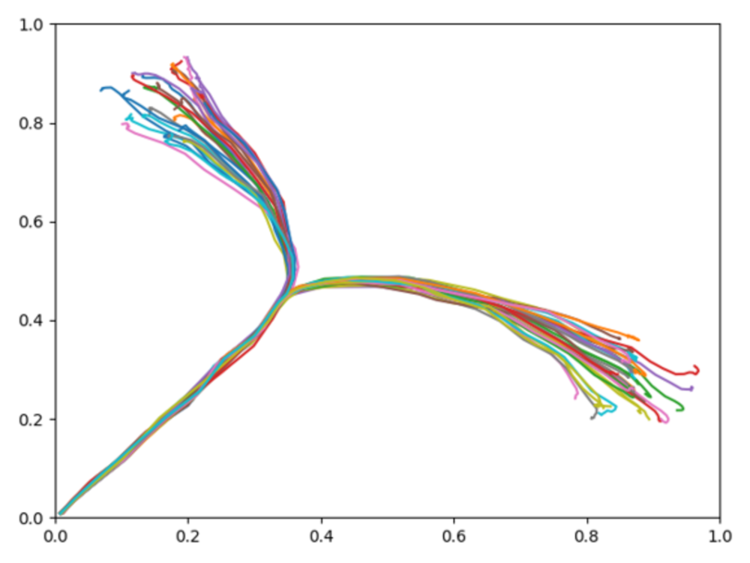}}
\hspace*{\fill}
\subcaptionbox{\label{fig:undstw}}{\includegraphics[width=0.3\textwidth]{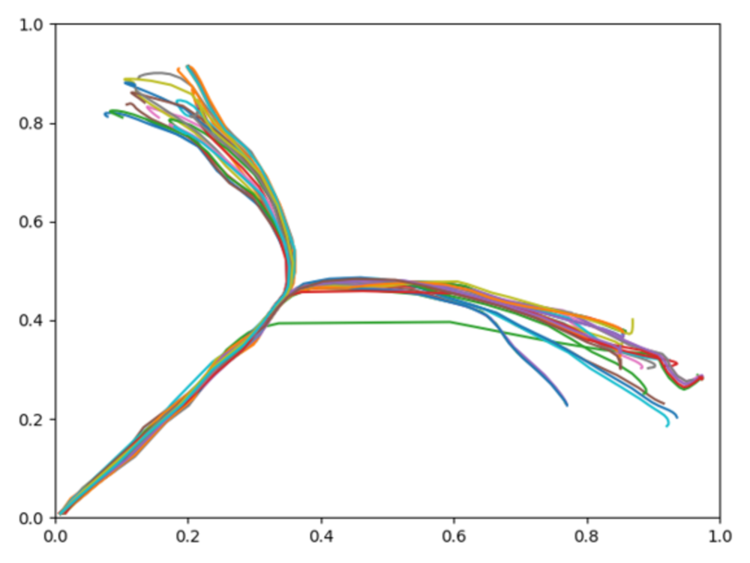}}
\hspace*{\fill}
\vspace{1em}
\caption{Trajectory plots showing \textbf{(a)} the training data (ground truth), \textbf{(b)} prior generation with a weak meta-prior, \textbf{(c)} with an intermediate meta-prior, and \textbf{(d)} with a strong meta-prior. Each plot contains 60 trajectories.\label{fig:priorgenw}}
\end{figure}

Table \ref{tbl:unddist} shows the distributions between left and right goal areas for the three networks trained with different meta-priors in comparison to the training data (ground truth). We observed that a weaker meta-prior tended to allow slightly more skew in one direction, however by inspecting the plots in Figure \ref{fig:priorgenw}, it is apparent that there is a large amount of noise in the trajectories generated by the weak meta-prior network (Figure \ref{fig:undwtw}). In particular, there are large deviations and the overall shape of the trajectories does not follow the training data accurately.

\begin{table}[!htbp]
\caption{Distribution of goals reached by networks with different meta-priors, after 60 prior generation sequences\label{tbl:unddist}}
\centering
\begin{tabular}{ccc}
\toprule
\textbf{Training meta-prior} & \textbf{Left goal \%} & \textbf{Right goal \%} \\
\midrule
Weak                         & 38.3                  & 61.7\\
Intermediate                 & 46.7                  & 53.3\\
Strong                       & 55.0                  & 45.0\\
\emph{Ground truth}          & \emph{50.0}           & \emph{50.0}\\
\bottomrule
\end{tabular}
\end{table}

In contrast, with a large meta-prior (Figure \ref{fig:undstw}), there appears to have been a failure to learn the spread of goals, particularly in the right goal area, resulting in some unexpected trajectories. In this test, an intermediate meta-prior was best at learning the probabilistic structure of the training data.

\subsubsection{Target regeneration}
As described previously, PV-RNN learns the probabilistic structure of trajectories by embedding in either initial state sensitive deterministic dynamics or stochastic dynamics based on the meta-prior value. This suggests that trajectories for goal-directed behaviors with multiple goals, including decision branching points, can be generated either in an initial state sensitive manner based on deterministic dynamics or in a noise-driven manner based on stochastic dynamics.

For the purpose of examining such properties of the trained networks, we conduct a test for target regeneration of the trained trajectories in a manner similar to that originally used in \cite{ahmadi19}. In this test, we attempt to regenerate a particular target sequence from the training dataset by using the information of the latent state in the initial step. This information was a result of the training process.

\begin{figure}[!htbp]
\centering
\includegraphics[width=0.5\textwidth]{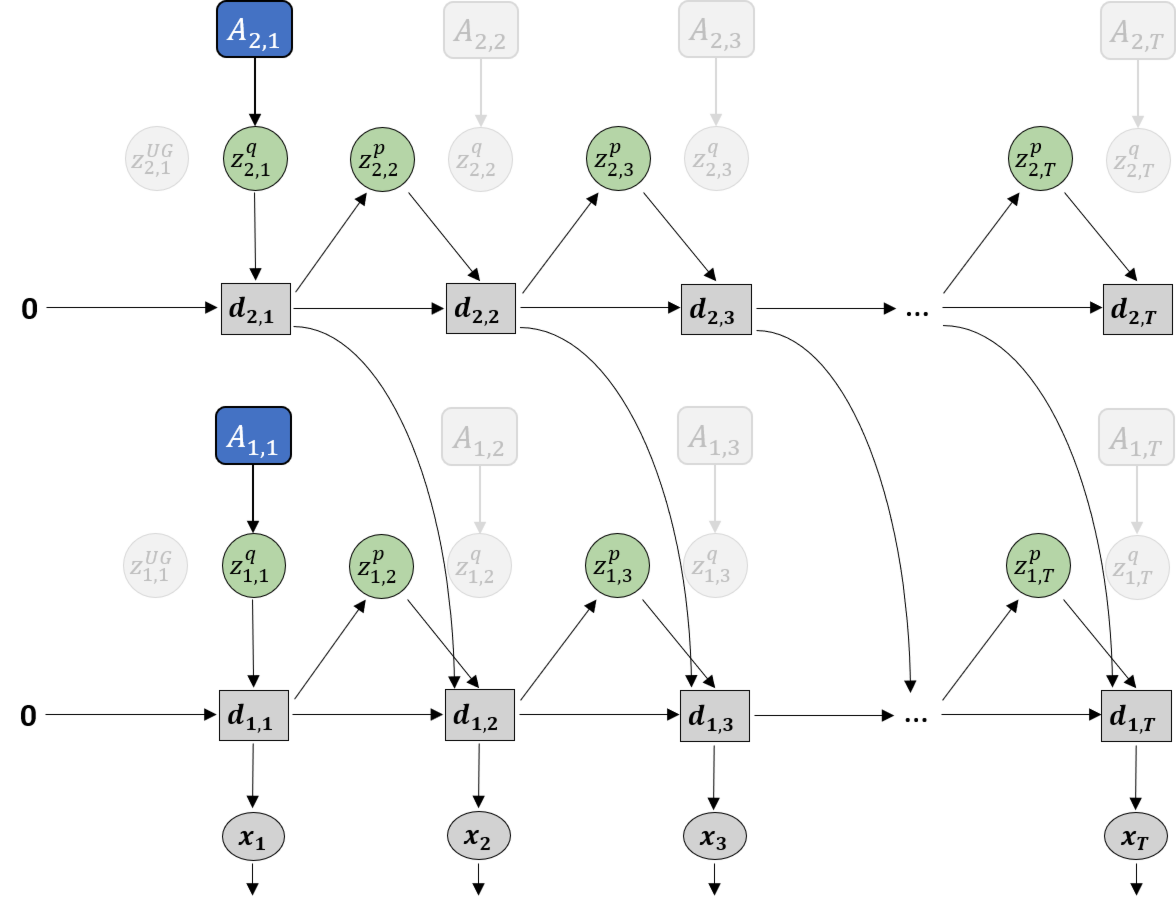}
\caption{Generation using a given posterior adaptation variable $\bm{A}_1$}
\label{fig:targetregen}
\end{figure}

More specifically, as illustrated in Figure \ref{fig:targetregen}, the prior generation is computed but with the posterior adaptation variable at $t=1$, $\bm{A}_1$, fixed to the value obtained after training on the sequence. This results in the setting of the initial latent state values of $Z_1$ and $d_1$ with values for the trained sequence.

\begin{figure}[!htbp]
\centering
\hspace*{\fill}
\subcaptionbox{\label{fig:tgttrain}}{\includegraphics[width=0.3\textwidth]{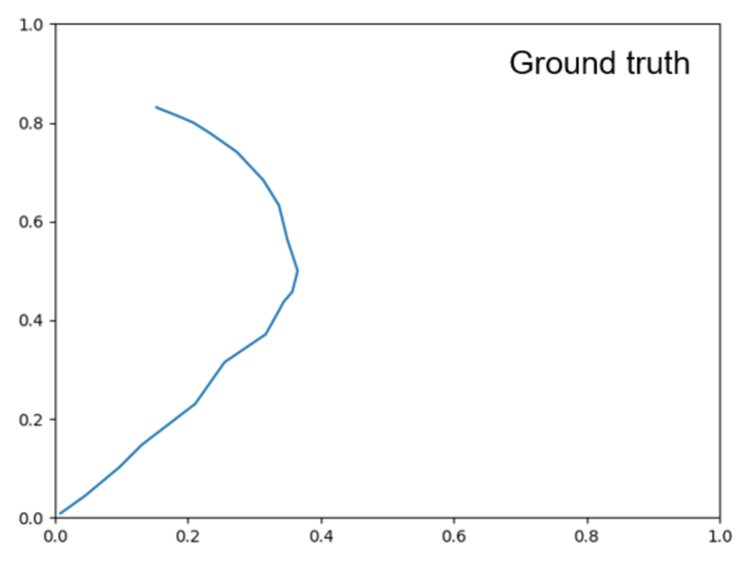}}
\hspace*{\fill}

\vspace{1em}
\hspace*{\fill}
\subcaptionbox{\label{fig:tgtwtw}}{\includegraphics[width=0.3\textwidth]{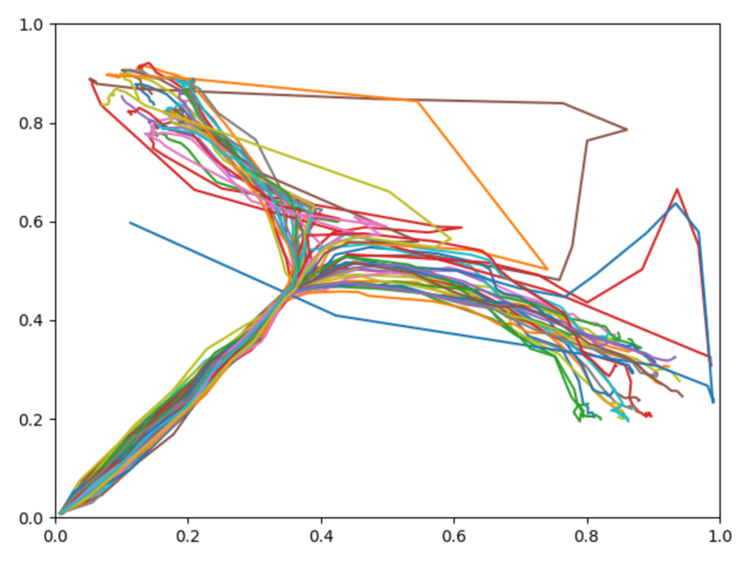}}
\hspace*{\fill}
\subcaptionbox{\label{fig:tgtitw}}{\includegraphics[width=0.3\textwidth]{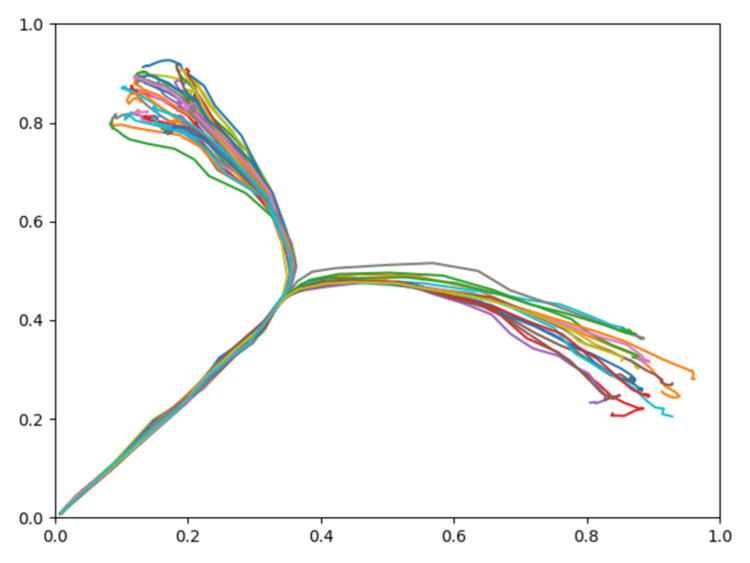}}
\hspace*{\fill}
\subcaptionbox{\label{fig:tgtstw}}{\includegraphics[width=0.3\textwidth]{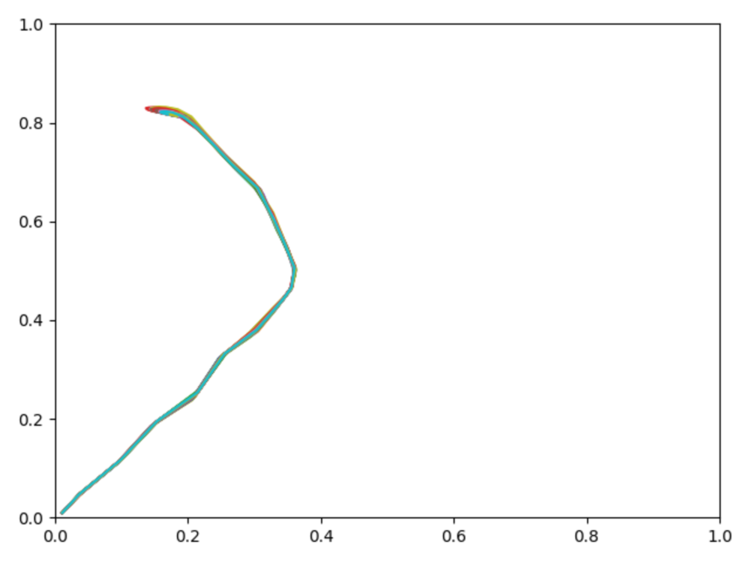}}
\hspace*{\fill}
\vspace{1em}
\caption{Trajectory plots showing \textbf{(a)} the target (ground truth), \textbf{(b)} target regeneration with a weak meta-prior, \textbf{(c)} target regeneration with an intermediate meta-prior, and \textbf{(d)} target regeneration with a strong meta-prior. Each plot contains 60 trajectories.\label{fig:tgtregenw}}
\end{figure}

In this test, we load a particular $\bm{A}_1$ adaptation value for a single training sequence that ends in the left goal (shown in Figure \ref{fig:tgttrain}) as an initial state, and allow the network to generate the trajectory 60 times from the same initial state with different noise sampling in the Z-units. If the information in the initial state is sufficient to regenerate the trajectory, and the network is deterministic, the generated trajectory should always match the ground truth. We refer to this condition as `initial state sensitive'.

\begin{table}[!htbp]
\centering
\caption{Distribution of goals reached by networks with different meta-priors, after 60 target regeneration sequences\label{tbl:tgtdist}}
\begin{tabular}{ccc}
\toprule
\textbf{Training meta-prior} & \textbf{Left goal \%} & \textbf{Right goal \%} \\
\midrule
Weak                         & 56.7                  & 43.3\\
Intermediate                 & 70.0                  & 30.0\\
Strong                       & 100.0                 & 0.0\\
\emph{Target}              & \emph{100.0}            & \emph{0.0}\\
\bottomrule
\end{tabular}
\end{table}

Results obtained after 60 target regeneration runs are summarized in Table \ref{tbl:tgtdist} and Figure \ref{fig:tgtregenw}. Table \ref{tbl:tgtdist} shows that the weaker meta-prior networks tend to ignore the target and retain the prior distribution seen previously. As the meta-prior increases, the distribution of goals reached skews towards the target. From this, we can surmise that a strong training meta-prior creates a network that is initial state sensitive, while networks with a weaker meta-prior do not show such a tendency. 

Visually inspecting the plots in Figure \ref{fig:tgtregenw} shows that in comparison to the prior generation results in Figure \ref{fig:priorgenw}, while the overall distribution of the output from the weak and intermediate meta-prior networks have not been affected by the initial state $\bm{A}_1$ the trajectories are not as stable. We also note that while a strong meta-prior produces a network with a strong initial state sensitivity, the result is not completely deterministic as there is still a noise component in the posterior distribution.

\begin{figure}[!htbp]
\centering
\hspace*{\fill}
\subcaptionbox{\label{fig:lz1}}{\includegraphics[width=0.3\textwidth]{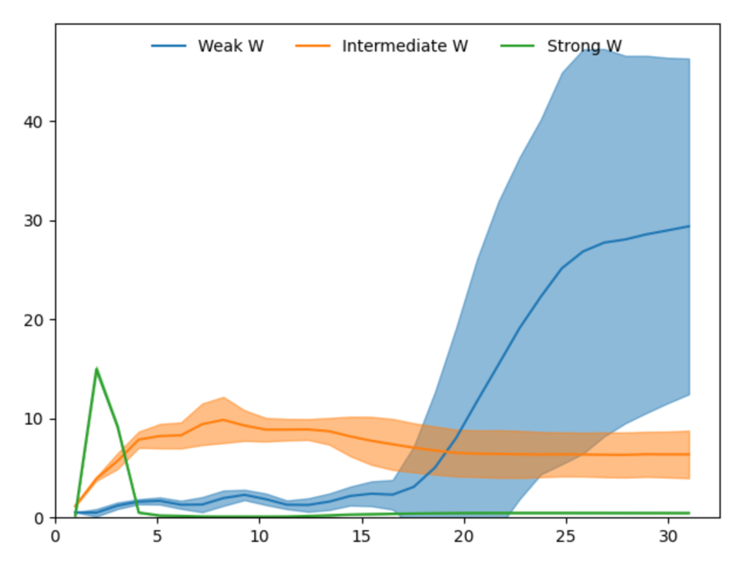}}
\hspace*{\fill}
\subcaptionbox{\label{fig:lz2}}{\includegraphics[width=0.3\textwidth]{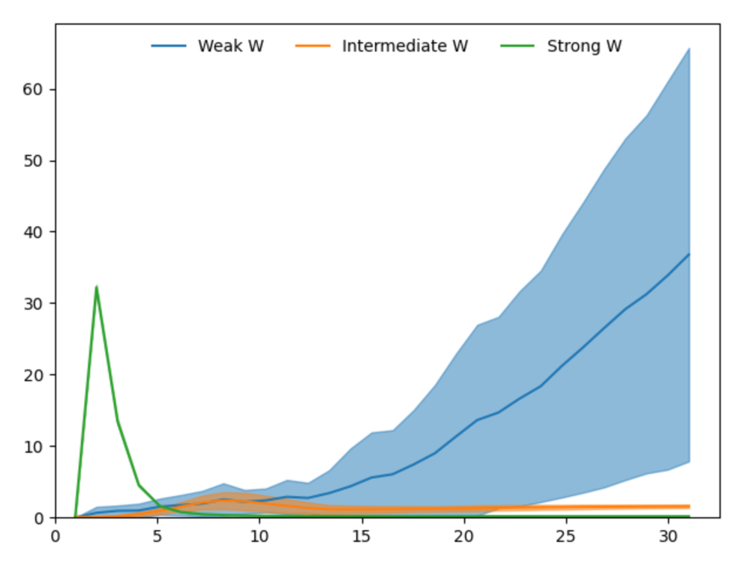}}
\hspace*{\fill}
\subcaptionbox{\label{fig:lz2plus}}{\includegraphics[width=0.3\textwidth]{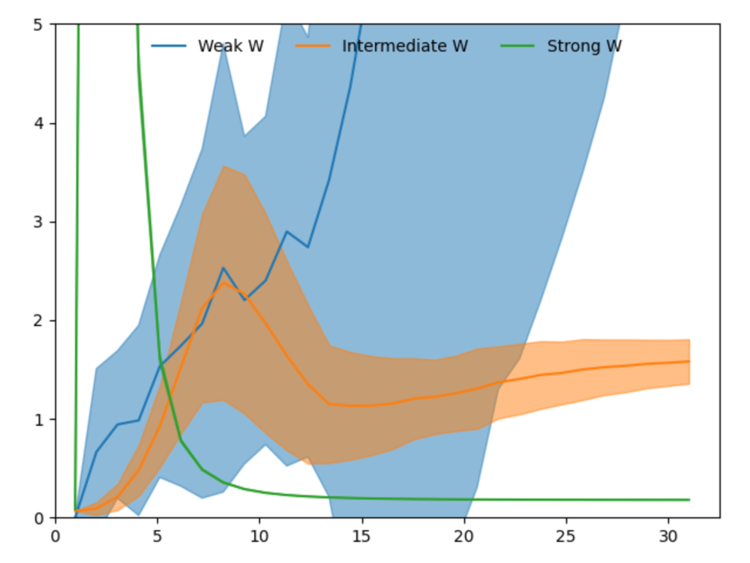}}
\hspace*{\fill}
\vspace{1em}
\caption{Plots of KLD during target regeneration given a particular $\bm{A}_1$ adaptation value. \textbf{(a)} Shows KLD for weak, intermediate and strong meta-prior in the bottom layer, \textbf{(b)} shows KLD for weak, intermediate and strong meta-prior in the top layer. \textbf{(c)} Adjusts the scale of (b) so the intermediate meta-prior result can be more clearly seen. The peak in KLD in the intermediate meta-prior network is visible around $t=8$. The shaded areas indicate the standard deviation of KLD over 60 generated trajectories.\label{fig:lz}}
\end{figure}

\begin{figure}[!htbp]
\centering
\hspace*{\fill}
\subcaptionbox{\label{fig:tgtwtwx}}{\includegraphics[width=0.3\textwidth]{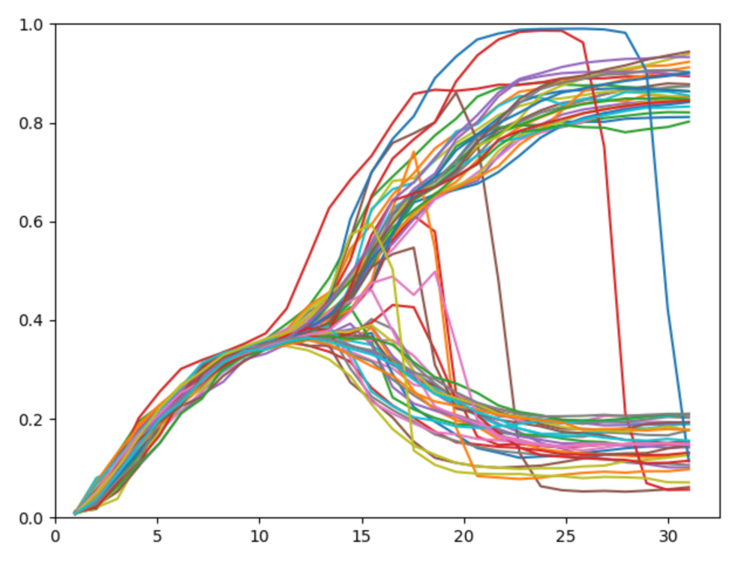}}
\hspace*{\fill}
\subcaptionbox{\label{fig:tgtitwx}}{\includegraphics[width=0.3\textwidth]{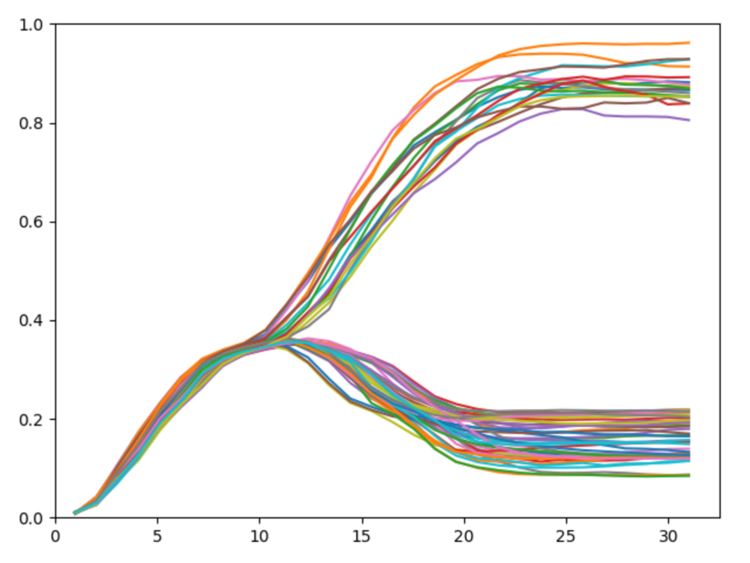}}
\hspace*{\fill}
\subcaptionbox{\label{fig:tgtstwx}}{\includegraphics[width=0.3\textwidth]{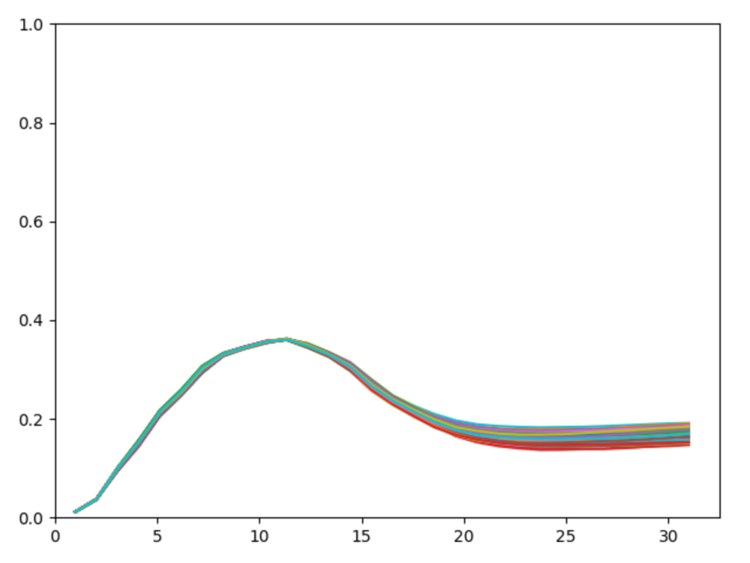}}
\hspace*{\fill}
\vspace{1em}
\caption{Plots of the $x$ coordinate over time, target regeneration given a particular $\bm{A}_1$ adaptation value with \textbf{(a)} a weak meta-prior, \textbf{(b)} an intermediate meta-prior, and \textbf{(c)} a strong meta-prior. The branch point is visible around $t=10$, except in \textbf{(c)} which does not exhibit any branching behavior. \label{fig:tgtregenwx}}
\end{figure}

To better illustrate the activity of the networks in regenerating a particular sequence given $\bm{A}_1$ at each timestep, Figure \ref{fig:lz} shows plots of the KLD at both layers while Figure \ref{fig:tgtregenwx} shows plots of the $x$ coordinates over time.

With a strong meta-prior, a large spike in KLD at $t=2$ followed by a rapid drop to near zero is visible at both layers, due to the posterior taking a particular mean with negligible variance to reconstruct the trajectory in an initial sensitive manner. Note that at $t=1$, KLD is regulated independently by $w_I$ so all the networks show identical behavior. This suggests the branching decision is taken early with a strong meta-prior in order to minimize average KLD.

The case with an intermediate meta-prior shows a more moderate peak in KLD around $t=8$ or $t=9$, just before the branch point at $t=10$, with KLD remaining flat toward the end. This suggests that uncertainty in the prior increases slowly until the branch point while uncertainty in the posterior is kept smaller in order to minimize the reconstruction error. Therefore, it is considered that the branch decision is built up by slowly accumulating sampled noise in the Z-units until the branch point. 

In Figure \ref{fig:tgtitwx}, we can clearly see the branch point at approximately $t=10$. The spread of the trajectories is relatively low until the branch, after which is a larger spread of trajectories until the goal points are reached.

With a weak meta-prior, KLD behaves slightly differently in the two layers. In the top layer, the KLD of the weak meta-prior network behaves similarly to the intermediate meta-prior network until the branch point, after which the KLD does not decline---rather it continues to rise and the spread of KLD values also increases significantly. In the bottom layer, KLD remains low until around $t=18$, where there is a sudden rise in KLD.

In this case, we surmise that uncertainty in the prior continues to build even after the branch point. This is visible in Figure \ref{fig:tgtwtwx} with a number of trajectories suddenly switching from one goal to the other after the branch point. Note that due to weighting of KLD by the weak meta-prior, the complexity term remains small, resulting in little pressure to follow learned priors. In addition, the high uncertainty toward the end of the sequences is likely the reason why in the following test the weak meta-prior network is able to generate trajectories that end closer to untrained goals than other networks.

\subsubsection{Plan generation} \label{sec:2Dplangen}
To evaluate GLean, we first took the three networks previously trained with different meta-priors and then ran plan generation to evaluate the impact of meta-prior during training on the generated motor plans. Plan generation was conducted using a test dataset containing 20 untrained sequences, with the initial XY and goal XY coordinates of each test sequence used to generate 20 motor plans. The goal coordinates of the test data are taken from the same distribution as in the training data. GLean was allowed to run for 500 epochs, with a plan adaptation rate (equivalent to learning rate in training) of 0.05. The generated motor plans were then compared against the ground truth sequences.

\begin{figure}[!htbp]
\centering
\hspace*{\fill}
\subcaptionbox{\label{fig:testdata}}{\includegraphics[width=0.3\textwidth]{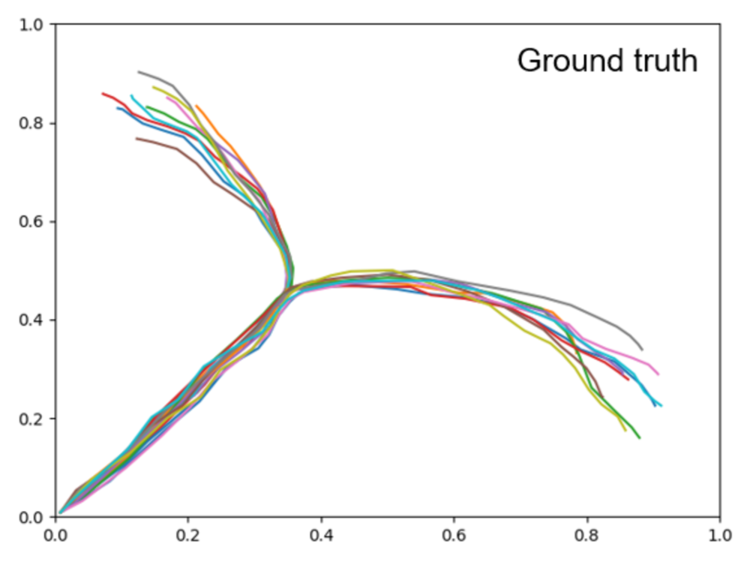}}
\hspace*{\fill}

\vspace{1em}
\centering
\hspace*{\fill}
\subcaptionbox{\label{fig:plan1ww}}{\includegraphics[width=0.3\textwidth]{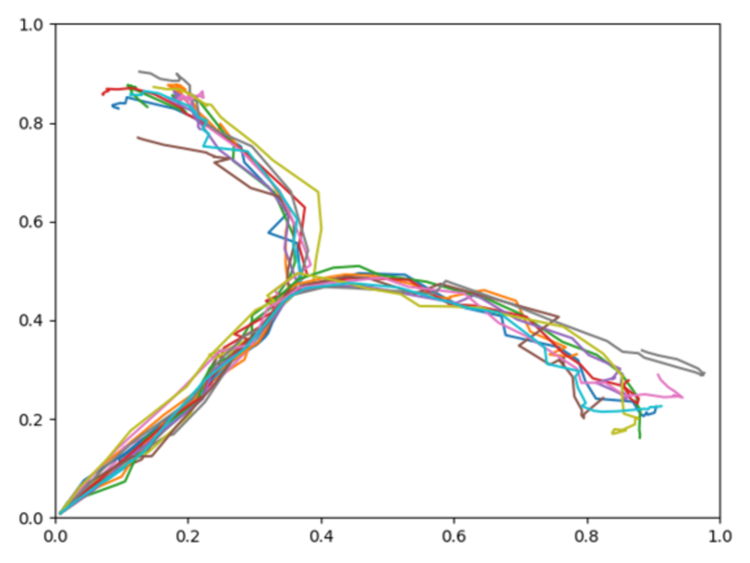}}
\hspace*{\fill}
\subcaptionbox{\label{fig:plan1ii}}{\includegraphics[width=0.3\textwidth]{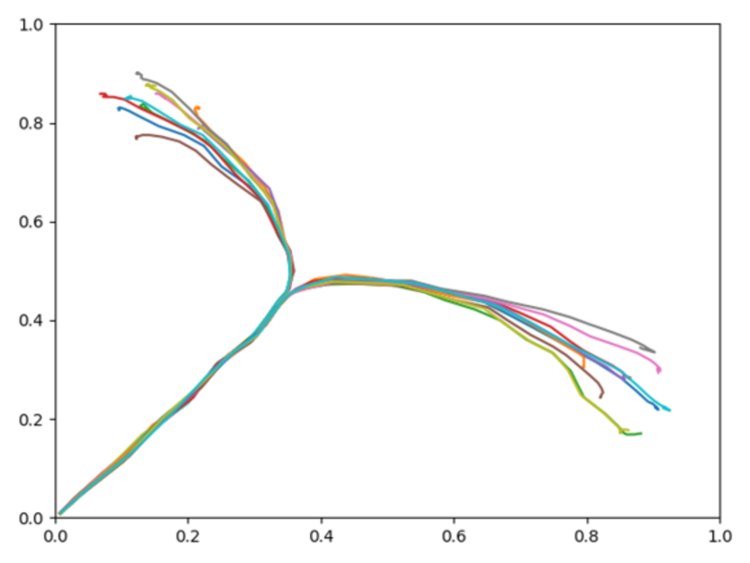}}
\hspace*{\fill}
\subcaptionbox{\label{fig:plan1ss}}{\includegraphics[width=0.3\textwidth]{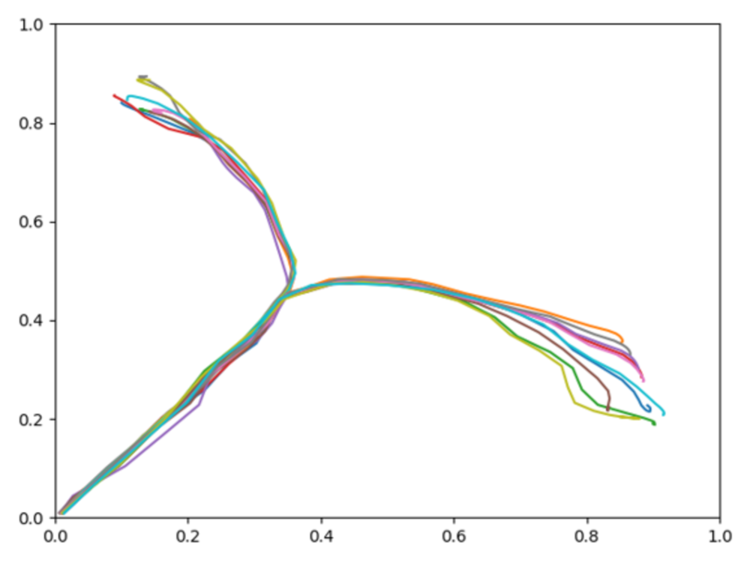}}
\hspace*{\fill}
\vspace{1em}
\caption{Plots showing motor plans of the 20 test sequences. \textbf{(a)} Shows the ground truth for untrained test data set, with the remaining plots generated with a \textbf{(b)} weak meta-prior, \textbf{(c)} intermediate meta-prior, and \textbf{(d)} strong meta-prior as described in Table \ref{tbl:exp1w}.\label{fig:plan1}}
\end{figure}

In the following results, we present quantitative results in a table, along with the meta-prior setting for training and planning. As plan generation using GLean is by necessity a non-deterministic process, plan generation was repeated 10 times for each network, with the result being averaged over the 10 runs. Average root mean squared error (RMSE) represents how closely the generated plans match the ground truth trajectories for each goal, while the average goal deviation (GD) represents the final distance to the goal. For both RMSE and GD, the standard deviation over 10 runs is given, and the lowest result is highlighted as the best. Average $KLD_{pq}$ represents the KL-divergence between the prior and posterior distributions, unweighted by the meta-prior. A low average $KLD_{pq}$ indicates the generated plans follow the learned prior distribution closely, while a high average $KLD_{pq}$ indicates significant deviation from learned patterns. Qualitative results are shown in trajectory plots that show all 20 generated trajectories and can be visually compared to the training data in Figure \ref{fig:testdata}.

\begin{table}[!htbp]
\centering
\caption{Plan generation results on the 20 trajectory test set with varying meta-prior. Best result highlighted in bold\label{tbl:plan1}}
\begin{tabular}{cccc}
\toprule
\textbf{Meta-prior} & \textbf{Average $\bm{KLD_{pq}}$} & \textbf{Average RMSE$\bm{\pm \sigma}$}    & \textbf{Average GD$\bm{\pm \sigma}$}\\
\midrule
Weak                & 159.1                            & $0.0615 \pm 0.0042$      & $\bm{5.2 \times 10^{-6} \pm 2.1 \times 10^{-6}}$\\
Intermediate        & 3.36                             & $\bm{0.0344 \pm 0.0021}$ & $7.8 \times 10^{-5} \pm 1.9 \times 10^{-5}$\\
Strong              & 0.17                             & $0.0375 \pm 0.0015$      & $6.7 \times 10^{-4} \pm 8.8 \times 10^{-5}$\\
\bottomrule
\end{tabular}
\end{table}

When conducting plan generation with the three networks trained with different meta-prior values, Table \ref{tbl:plan1} shows that while the intermediate meta-prior network has the lowest average RMSE, the weak meta-prior network has the lowest goal deviation. As expected, the average KLD increases as the meta-prior weight is reduced, although we note that while the increase in KLD between strong and intermediate is inversely proportional to the change in meta-prior, the average KLD increase from intermediate to weak does not follow the same relationship. In addition, while it appears that the weak meta-prior trades a factor of 2 reduction of average RMSE for a factor of 10 improvement in goal deviation, the plots in Figure \ref{fig:plan1} show that the trajectories generated by the weak meta-prior network are very noisy compared to the output from the other two networks.

From visual inspection of the trajectory plots compared to the ground truth (Figure \ref{fig:plan1}), we can additionally see that the intermediate meta-prior network tends to average the trajectories more than the strong meta-prior network that is following the training data more strongly and as a result tends to miss the goal. 
In summary, plans were generated with highest generalization in the case of an intermediate value for meta-prior, whereas the planned trajectories became significantly more noisy in the case with a weak meta-prior, and in the case with a strong meta-prior the trajectories could not reach the specified goals well. In Section \ref{sec:discussion}, we discuss further the implications of the meta-prior setting in plan generation.

\subsubsection{Plan generation for goals set in unlearned regions}
\begin{figure}[!htbp]
\centering
\hspace*{\fill}
\subcaptionbox{\label{fig:centergt}}{\includegraphics[width=0.3\textwidth]{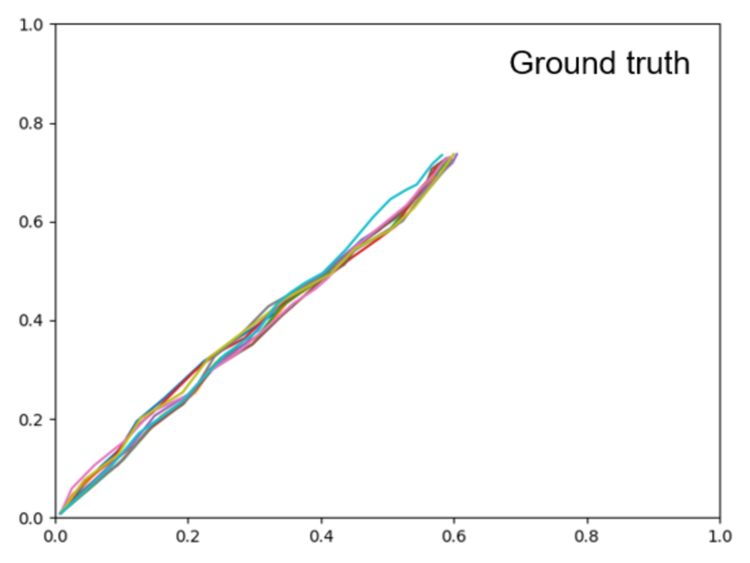}}
\hspace*{\fill}
\subcaptionbox{\label{fig:centerplan}}{\includegraphics[width=0.3\textwidth]{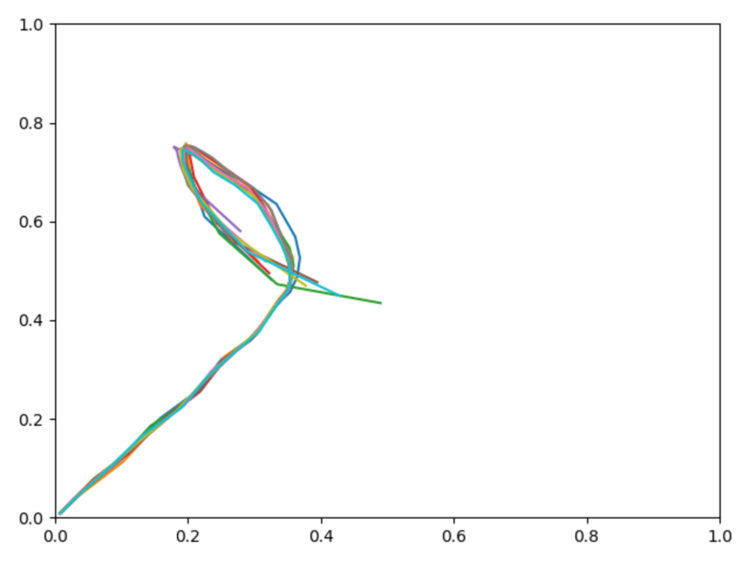}}
\hspace*{\fill}
\vspace{1em}
\caption{Plots showing motor plans of the 10 test sequences with goals set in an untrained region. \textbf{(a)} Shows the ground truth test trajectories, and \textbf{(b)} shows the results of plan generation.\label{fig:center1}}
\end{figure}

The following test examines whether GLean can achieve goals set outside of the trained regions. Figure \ref{fig:centergt} shows 10 ground truth trajectories reaching goals in an untrained region which is in the middle of the left and right trained goal regions.

Figure \ref{fig:centerplan} shows the results of plan generation performed by the network trained with the intermediate meta-prior, where it is apparent that GLean is not able to reach the specified goals. In particular, it can be observed that the trajectories cannot go straight at the branching point. Instead, they branch left and head towards the left goal region. This is likely because the learned prior strongly prefers either turning left or right but not going straight at the branching point. This result implies that GLean is more likely to generate goal-directed plan trajectories within well habituated areas.

\subsection{Experiment 2: simulated robotic object manipulation task} \label{sec:robotexp}
\begin{figure}[!htbp]
\setlength{\fboxsep}{0pt}
\centering
\hspace*{\fill}
\subcaptionbox*{}{\fbox{\includegraphics[width=0.3\textwidth]{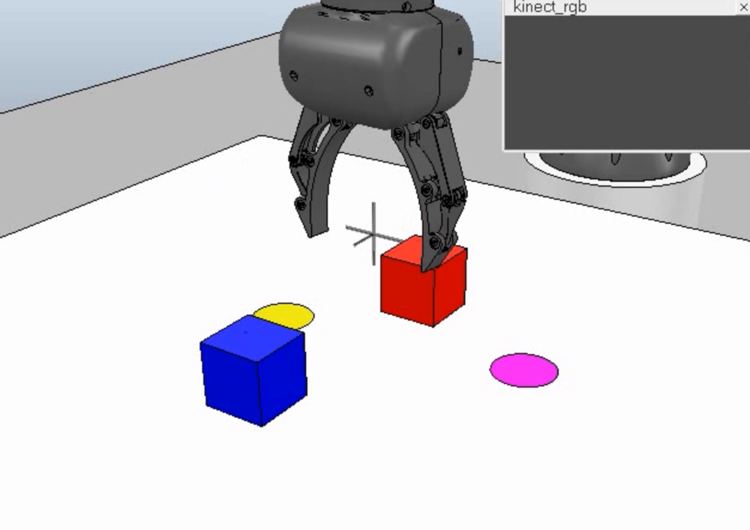}}}
\hspace*{\fill}
\subcaptionbox*{}{\fbox{\includegraphics[width=0.3\textwidth]{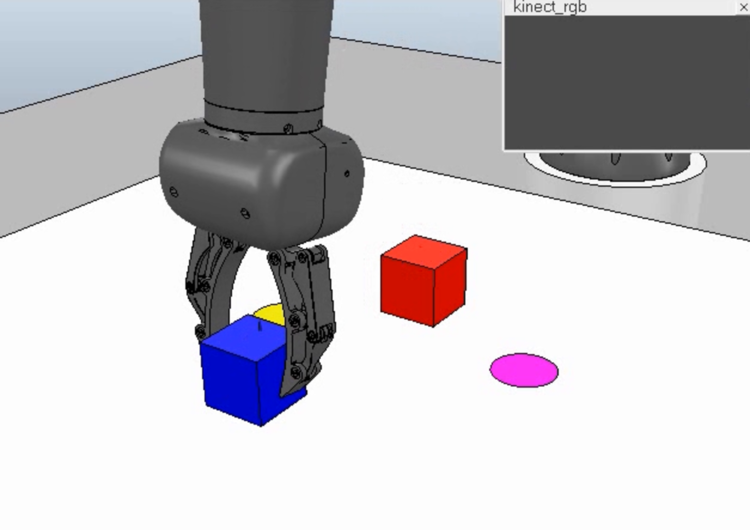}}}
\hspace*{\fill}
\subcaptionbox*{}{\fbox{\includegraphics[width=0.3\textwidth]{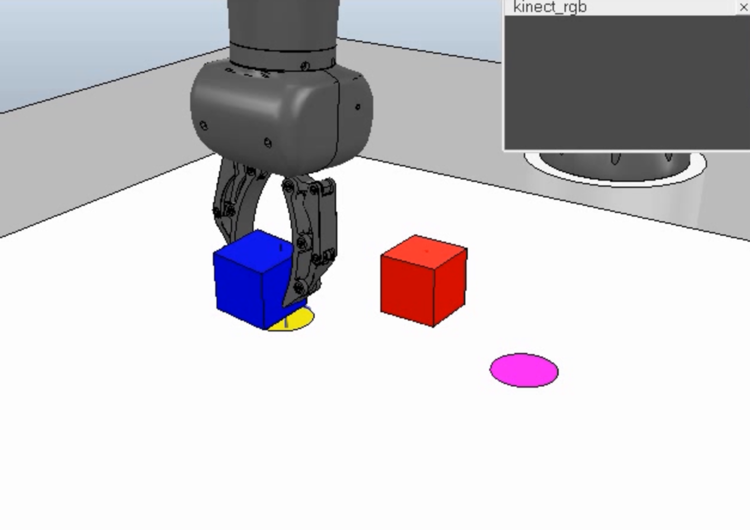}}}
\hspace*{\fill}
\caption{Simulated robot executing the grasp and place task. In the workspace in front of the robot, there are two graspable blocks and two goal circles. Crosshair markers show the predicted positions of the gripper and the two blocks.\label{fig:exp2}}
\end{figure}

In order to test the performance of GLean in a robotic environment, we prepared a simulated environment in the V-REP simulator with a model of a real 8 DOF arm robot (a Tokyo Robotics Torobo Arm) with a gripper end effector (see Figure \ref{fig:exp2}). In front of the robot is a workspace of approximately 30cm square, with two cube blocks (5cm/side) and two circles (5cm diameter). The robot always starts in a set home position with the gripper between and above the four objects, the blocks placed in front and behind the gripper, and the circles placed left and right of the gripper. The positions of the objects are randomized following a Gaussian distribution, with $\sigma \simeq 3$cm. The task is to generate a motor plan to grasp one of the blocks and place it on a disc, as well as predict the coordinates of the gripper and the two blocks.

\begin{figure}[!htbp]
\centering
\includegraphics[width=0.5\textwidth]{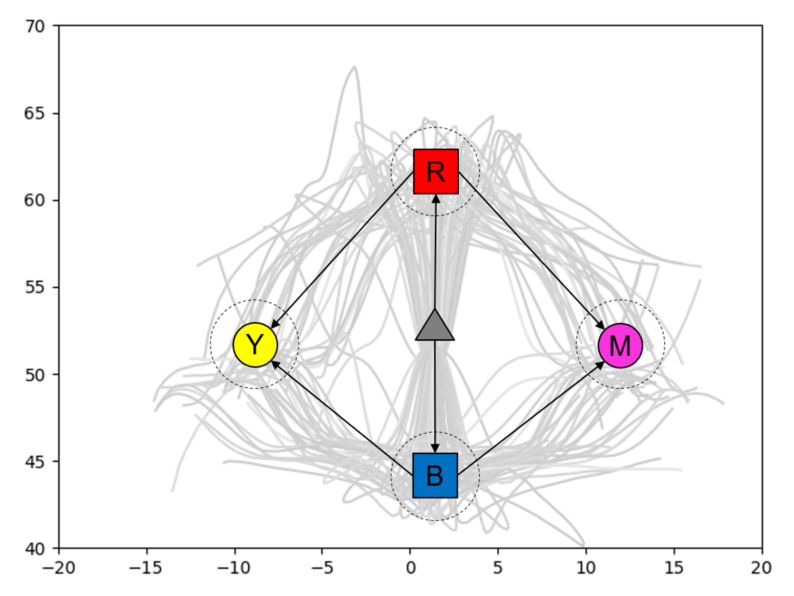}
\caption{Trajectories of the gripper in two dimensions, with the mean positions of the blocks and goal circles overlaid. Dashed circles represent the standard deviation of the positions. \label{fig:exp2gripperxy}}
\end{figure}

In order to achieve this, the robot is trained with 120 trajectories that involve the robot arm (1) moving forward or backward to grasp the appropriate object, then (2) carry the object to the desired goal. Figure \ref{fig:exp2gripperxy} shows the trajectories of the gripper in two dimensions, overlaid with an illustration of the gripper and objects. The training data is generated by a custom kinematic routine that converts the workspace coordinates to a series of pre-recorded movements taken from our previous work with this robot. The recorded trajectories are resampled to 80 timesteps, with padding at the end as necessary. 

The initial state of the environment is given as the joint angles of the robot as well as the 3D $(x, y, z)$ coordinates of the gripper and blocks (totaling 17 dimensions), while the goal is given as only the coordinates of the gripper and blocks. Using GLean, a motor plan to grasp the appropriate block and take it to the correct goal is generated, along with predictions of the coordinates of the gripper and both blocks at each timestep. At the end of the generated sequence, the robot releases the block and the control program records the distance between the centers of the block and goal circle. If the block and goal circle overlap, the trial is considered successful.

\begin{figure}[!htbp]
\centering
\hspace*{\fill}
\subcaptionbox{\label{fig:FW}}{\includegraphics[width=0.4\textwidth]{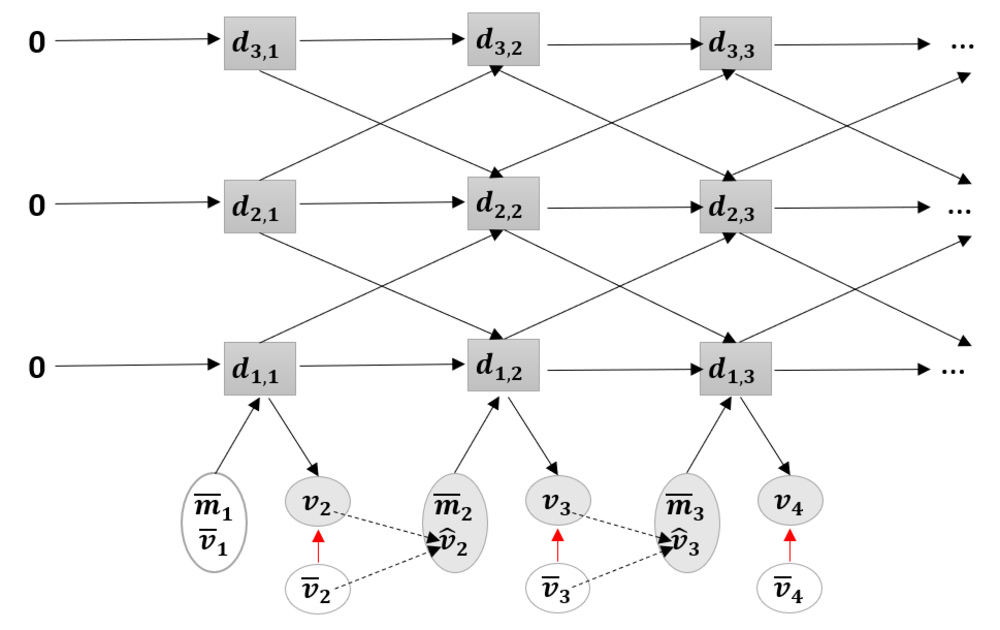}}
\hspace*{\fill}
\subcaptionbox{\label{fig:SI}}{\includegraphics[width=0.4\textwidth]{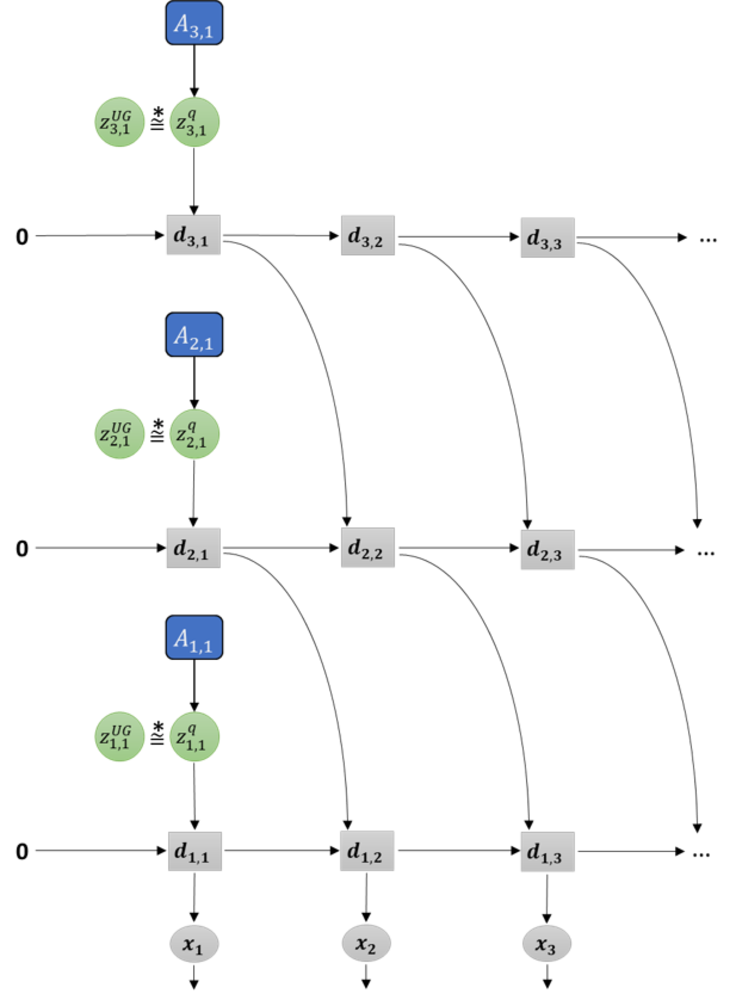}}
\hspace*{\fill}
\vspace{1em}
\caption{Graphical representations of \textbf{(a)} the forward model (FM) and \textbf{(b)} the stochastic initial state (SI) model as implemented in this paper}
\end{figure}

In order to compare the performance of GLean against other typical trajectory planning approaches, we have implemented a forward model with MTRNN (see Figure \ref{fig:FW}) and a stochastic initial state MTRNN (see Figure \ref{fig:SI}). FM, as mentioned previously, is commonly used in robotic trajectory planning and is able to predict the next sensory state given the current sensory and motor states. For this paper, it is implemented using the same MTRNN as used by GLean, but with no stochastic units. SI is implemented similarly, although using a 1:1 ratio of deterministic and stochastic units at $t=1$ and no stochastic units at $t>1$ (as in \cite{jung19}). Table \ref{tbl:exp2rnn} shows an overview of the network parameters for GLean, FM, and SI.

\begin{table}[!htbp]
\caption{Network parameters used for the simulated robot experiment for 3 different models---GLean, FM, and SI.} \label{tbl:exp2rnn}
\hspace*{\fill}
\begin{subtable}{0.3\textwidth}
\caption{GLean}
\centering
\begin{tabular}{cccc|}
\toprule
                            & \multicolumn{3}{c|}{MTRNN layer} \\
	                        & \textbf{1}	& \textbf{2}  & \textbf{3} \\
\midrule
Neurons $|\bm{d}^l|$	    & 30            & 20		  & 10 \\
Z-units	$|\bm{z}^l|$        & 3             & 2 		  & 1 \\
$\tau$                      & 2             & 4           & 8 \\
\bottomrule
\multicolumn{4}{l}{} \\
\end{tabular}
\end{subtable}
\hspace*{\fill}
\begin{subtable}{0.3\textwidth}
\caption{FM}
\centering
\begin{tabular}{cccc|}
\toprule
                            & \multicolumn{3}{c|}{MTRNN layer} \\
	                        & \textbf{1}	& \textbf{2}  & \textbf{3} \\
\midrule
Neurons $|\bm{d}^l|$	    & 30            & 20		  & 10 \\
Z-units	$|\bm{z}^l|$        & 0             & 0 		  & 0 \\
$\tau$                      & 2             & 4           & 8 \\
\bottomrule
\multicolumn{4}{l}{} \\
\end{tabular}
\end{subtable}
\hspace*{\fill}
\begin{subtable}{0.3\textwidth}
\caption{SI}
\centering
\begin{tabular}{cccc}
\toprule
                            & \multicolumn{3}{c}{MTRNN layer} \\
	                        & \textbf{1}	& \textbf{2}  & \textbf{3} \\
\midrule
Neurons $|\bm{d}^l|$	    & 30            & 20		  & 10 \\
Z-units	$|\bm{z}^l|$        & 30*           & 20*         & 10* \\
$\tau$                      & 2             & 4           & 8 \\
\bottomrule
\multicolumn{4}{l}{*Stochastic Z-units only at $t=1$} \\
\end{tabular}
\end{subtable}
\hspace*{\fill}
\end{table}

Both FM and SI are trained in a partially closed loop manner to improve their generative ability, blending the ground truth sensory state $\bm{\overline{v}_t}$ and the predicted sensory state $\bm{v_t}$ as in \cite{jung19}. For SI, we use global norm gradient clipping with a setting of 50 to ensure the network remains stable. Note that while the forward model has thus far been depicted as operating on the motor space directly, in this comparison the forward model operates in proprioception space to match the other models.

\begin{equation}
    \bm{\hat{v}_t} = 0.9 \bm{v_t} + 0.1 \bm{\overline{v}_t}
\end{equation}

For this experiment, we have adjusted the meta-prior settings as shown in Table \ref{tbl:exp2w}. The range of meta-prior values has been reduced significantly as this task is much more sensitive to this setting, as shown in the following results.

\begin{table}[!htbp]
\caption{Meta-prior settings for the simulated robot experiment} \label{tbl:exp2w}
\centering
\begin{tabular}{cccc}
\toprule
                                & \multicolumn{3}{c}{MTRNN layer} \\
\textbf{Meta-prior setting $w$} & \textbf{L1}	& \textbf{L2}   & \textbf{L2} \\
\midrule
Weak		                    & 0.0004        & 0.0002		& 0.0001 \\
Intermediate                    & 0.0008 		& 0.0004        & 0.0002 \\
Strong                          & 0.002         & 0.001         & 0.0005 \\
\bottomrule
\end{tabular}
\end{table}

\subsubsection{Plan generation}
As in plan generation results with the 2D dataset in Section \ref{sec:2Dplangen}, here we use a test set of 20 untrained trajectories, with the results being averaged over 10 plan generation runs. GLean was allowed to run for 1000 epochs, with a plan adaptation rate of 0.1. In Table \ref{tbl:exp2}, we compare generated trajectories to ground truth trajectories, and here it can be surmised again that finding an intermediate setting of the meta-prior gives the best result---not only in terms of being close to the ground truth but in terms of success rate in accomplishing the grasp and place task in the simulator. 

In Table \ref{tbl:exp2sim}, we summarize the results of executing the generated plans using the robot simulator---comprising of success rate at the task as well as the average distance of the final block position from the goal, the latter only being counted in successful attempts. As in our previous work with a similar grasping task, succeeding in this task requires high accuracy at the grasp point in the middle of the trajectory. Thus, even though the differences in the generated trajectories are relatively small, the outcomes in simulation are significantly altered.

Note that despite the weak meta-prior offering a theoretically lower goal deviation, the actual measured error at the goal is higher than the intermediate meta-prior network. This is due to the average distance from the goal measuring from the block center to the goal center, and if the block was off-center when it was grasped, it would likely be off-center when it is placed on the goal. Due to the weak meta-prior network being less accurate during the intermediate steps, higher errors at the grasp point is likely.

\begin{table}[!htbp]
\centering
\caption{GLean generated plans with networks trained with different meta-priors, compared with ground truth. Note that in order for the results in the following tables to be comparable to the previous experiment, the output values were rescaled to $[0,1]$. Only the sensory states are compared between generated and ground truth trajectories. Best result highlighted in bold\label{tbl:exp2}}
\begin{tabular}{cccc}
\toprule
\textbf{Meta-prior} & \textbf{Average $\bm{KLD_{pq}}$} & \textbf{Average RMSE$\pm \sigma$} & \textbf{Average GD$\pm \sigma$}\\
\midrule
Weak                & 12.48                            & $0.0387 \pm 0.00067$              & $\bm{5.7 \times 10^{-5} \pm 7.4\times10^{-6}}$\\
Intermediate        & 4.64                             & $\bm{0.0230 \pm 0.00058}$         & $6.9\times 10^{-5} \pm 8.6\times10^{-6}$\\
Strong              & 2.35                             & $0.0242 \pm 0.00051$              & $1.3 \times 10^{-4} \pm 1.4 \times 10 ^{-5}$\\
\bottomrule
\end{tabular}
\end{table}

\begin{table}[!htbp]
\centering
\caption{Simulation results of executing GLean generated plans with networks trained with different meta-priors. Best result highlighted in bold\label{tbl:exp2sim}}
\begin{tabular}{ccc}
\toprule
\textbf{Meta-prior} & \textbf{Success rate} & \textbf{Average error at goal$\pm \sigma$}\\
\midrule
Weak                & 51.5\%                & $1.74 \pm 0.15$cm\\
\textbf{Intermediate} & \textbf{86.0\%}     & $\bm{1.52 \pm 0.07}$\textbf{cm}\\
Strong              & 60.5\%                & $2.02 \pm 0.11$cm\\
\bottomrule
\end{tabular}
\end{table}

\begin{figure}[!htbp]
\centering
\hspace*{\fill}
\subcaptionbox*{\label{fig:exp2pgw1v}}{\includegraphics[width=0.24\textwidth]{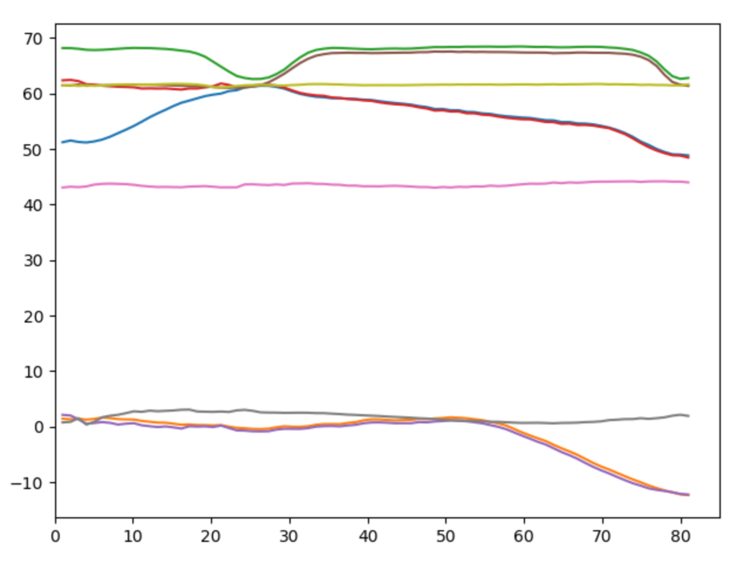}}
\hspace*{\fill}
\subcaptionbox*{\label{fig:exp2pgw3v}}{\includegraphics[width=0.24\textwidth]{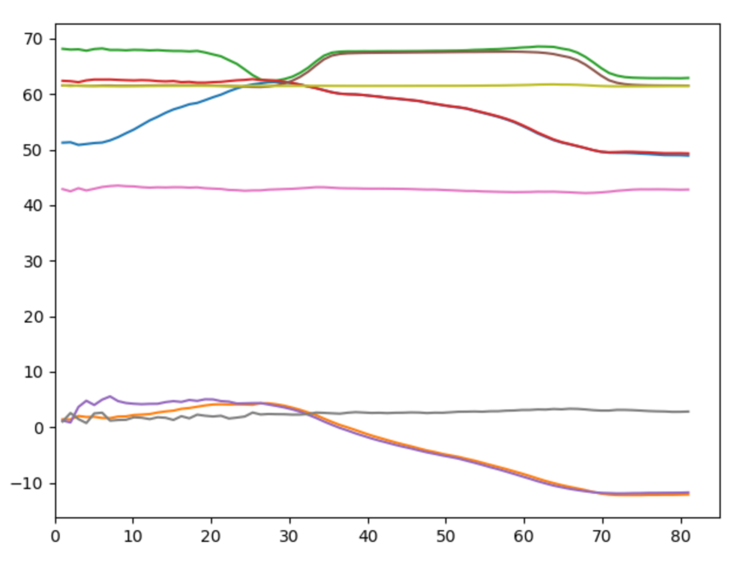}}
\hspace*{\fill}
\subcaptionbox*{\label{fig:exp2pgw7v}}{\includegraphics[width=0.24\textwidth]{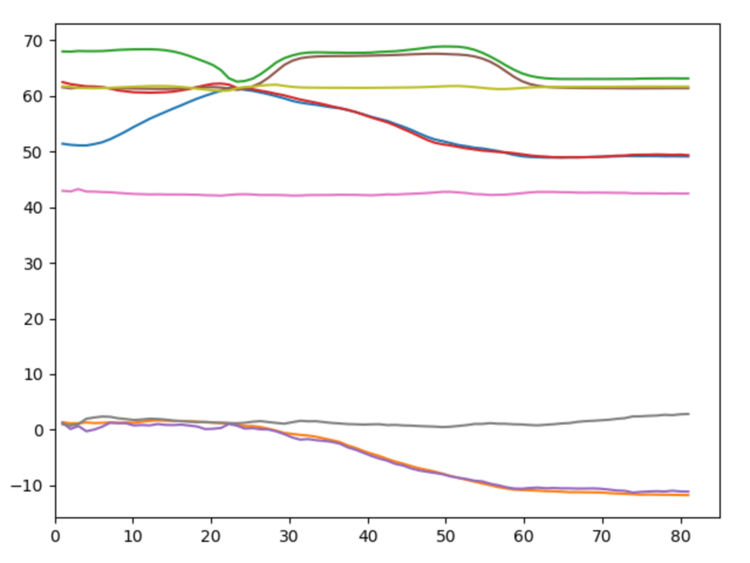}}
\hspace*{\fill}
\subcaptionbox*{\label{fig:exp2pggtv}}{\includegraphics[width=0.24\textwidth]{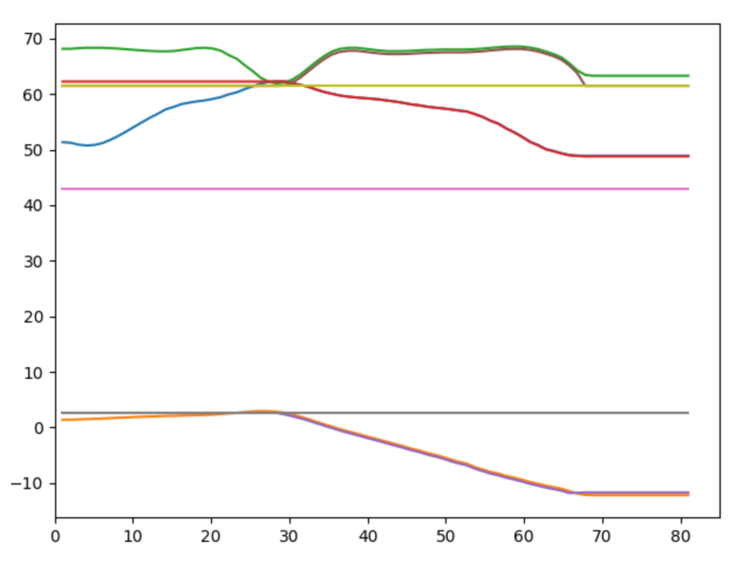}}
\hspace*{\fill}
\subcaptionbox{\label{fig:exp2pgw1m}}{\includegraphics[width=0.24\textwidth]{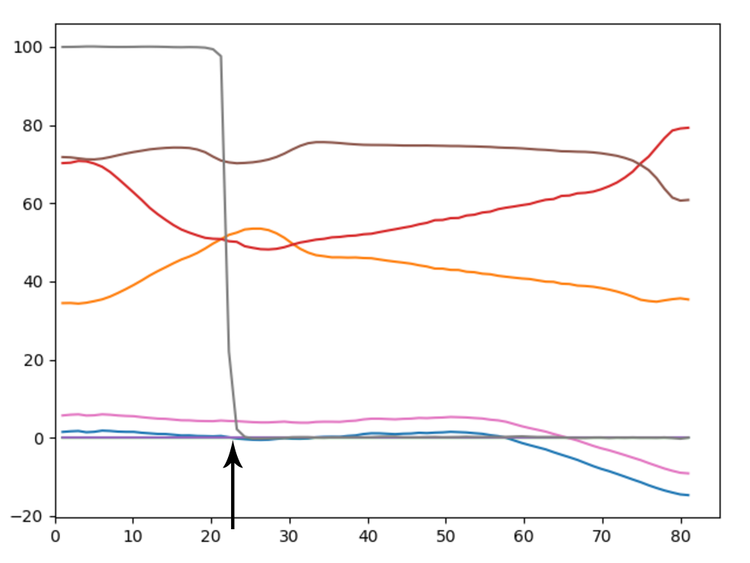}}
\hspace*{\fill}
\subcaptionbox{\label{fig:exp2pgw3m}}{\includegraphics[width=0.24\textwidth]{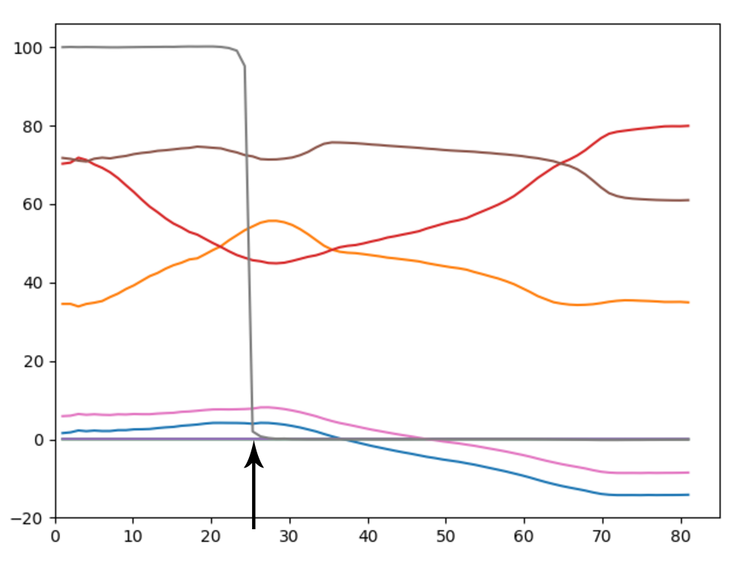}}
\hspace*{\fill}
\subcaptionbox{\label{fig:exp2pgw7m}}{\includegraphics[width=0.24\textwidth]{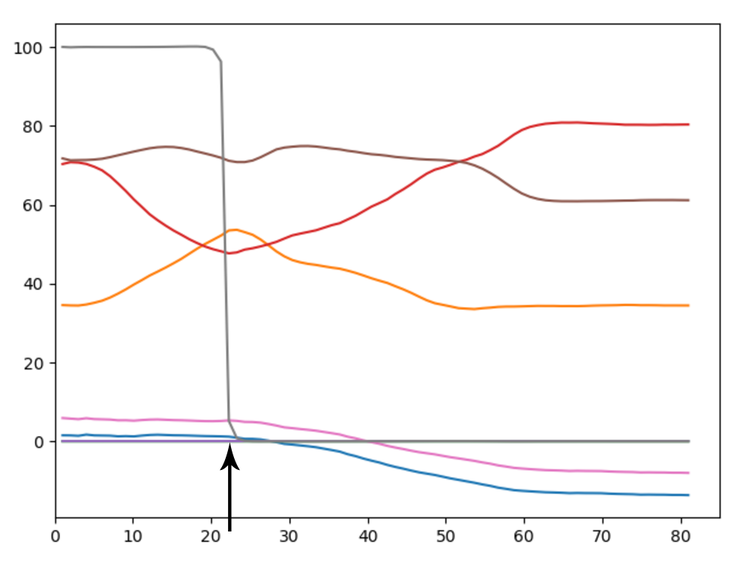}}
\hspace*{\fill}
\subcaptionbox{\label{fig:exp2pggtm}}{\includegraphics[width=0.24\textwidth]{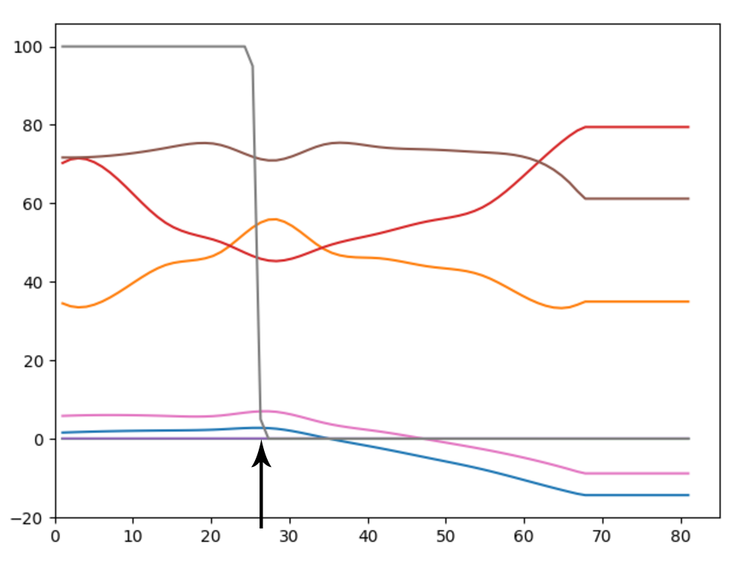}}
\hspace*{\fill}
\vspace{1em}
\caption{Plots showing the predicted sensory states (top row) and the motor plans (bottom row) for a given goal. The colored lines within each plot represent a sequence of predictions for one sensory or proprioception dimension. The columns of plots correspond to \textbf{(a)} weak meta-prior, \textbf{(b)} intermediate meta-prior, \textbf{(c)} strong meta-prior, and \textbf{(d)} ground truth. An arrow indicates the grasp point, where the robot attempts to pick up the block. While the exact timestep of the grasp point can vary, if the relationship between the predicted dimensions is not maintained the grasping attempt is more likely to fail. \label{fig:exp2plangen}}
\end{figure}

As an illustrative example, Figure \ref{fig:exp2plangen} shows generated plans consisting of predicted sensory and motor states given the initial environmental state and the goal sensory image. As suggested by the overall results, GLean trained with an intermediate meta-prior appears to generate trajectories that most resemble the ground truth trajectory. While in some other tasks it is possible that the trajectory between the start and the goal is not critical, in order for the robot to successfully complete the task in this experiment accuracy is required at the point where the robot grasps the block.

\subsection{Comparison between GLean, FM, and SI}
Finally, we compare the plan generation performance of GLean against the stochastic initial state (SI) model and the forward model (FM). GLean in this test is represented by the intermediate meta-prior network from the previous test. Results from SI are averaged over 10 runs, as with GLean. FM is deterministic and thus some statistics such as $KLD_{pq}$ and $\sigma$ are omitted. 

\begin{table}[!htbp]
\centering
\caption{Plan generation results of GLean, FM, and SI. Best result highlighted in bold\label{tbl:exp2cmp}}
\begin{tabular}{cccc}
\toprule
\textbf{Model}              & \textbf{Average $\bm{KLD_{pq}}$}  & \textbf{Average RMSE$\pm \sigma$} & \textbf{Average GD$\pm \sigma$}\\
\midrule
Forward model               & --                                & $0.1504$      & $6.8 \times 10^{-3}$\\
Stochastic initial state    & 3.32                              & $0.0257 \pm 0.00085$      & $7.8 \times 10^{-5} \pm 3.3\times10^{-6}$\\
\textbf{GLean}              & 4.64                              & $\bm{0.0230 \pm 0.00058}$ & $\bm{6.9\times 10^{-5} \pm 8.6\times10^{-6}}$\\
\bottomrule
\end{tabular}
\end{table}

From Table \ref{tbl:exp2cmp}, which evaluates the three algorithms' planning performance compared to the ground truth, we can observe that FM is the worst performer, with RMSE an order of magnitude and GD two orders of magnitude worse than either GLean or SI. On the other hand, GLean and SI are relatively close in this theoretical planning performance. However, as summarized in Table \ref{tbl:exp2cmpsim}, executing the generated plans in the robot simulator demonstrates a significant advantage for GLean. FM is unable to generate any plausible motor plans and thus achieved no successful runs.

\begin{table}[!htbp]
\centering
\caption{Simulation results of executing plans generated by GLean, FM, and SI. Best result highlighted in bold\label{tbl:exp2cmpsim}}
\begin{tabular}{ccc}
\toprule
\textbf{Model}                & \textbf{Success rate} & \textbf{Average error at goal$\pm \sigma$}\\
\midrule
Forward model (FM)            & 0.0\%                 & --\\
Stochastic initial state (SI) & 68.0\%                & $2.02 \pm 0.14$cm\\
\textbf{GLean}                & \textbf{86.0\%}       & $\bm{1.52 \pm 0.07}$\textbf{cm}\\

\bottomrule
\end{tabular}
\end{table}

\begin{figure}[!htbp]
\setlength{\fboxsep}{0pt}
\centering
\hspace*{\fill}
\subcaptionbox{FM\label{fig:exp2fwngt}}{\includegraphics[width=0.3\textwidth]{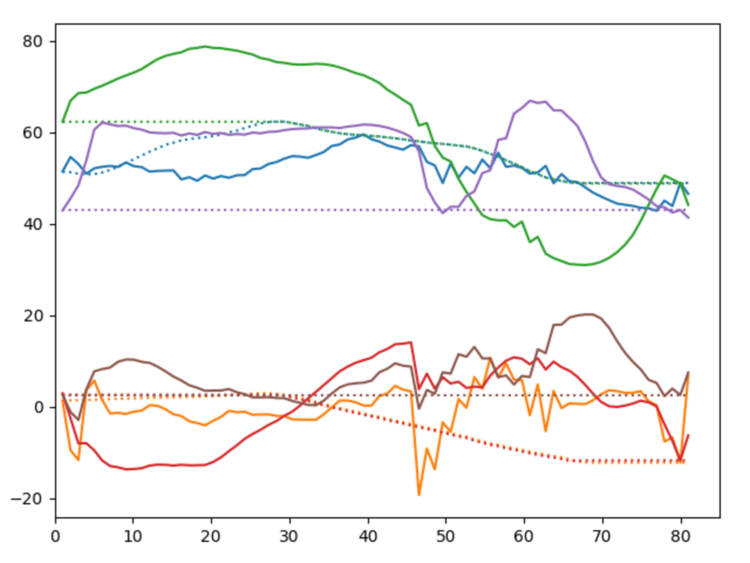}}
\hspace*{\fill}
\subcaptionbox{SI\label{fig:exp2singt}}{\includegraphics[width=0.3\textwidth]{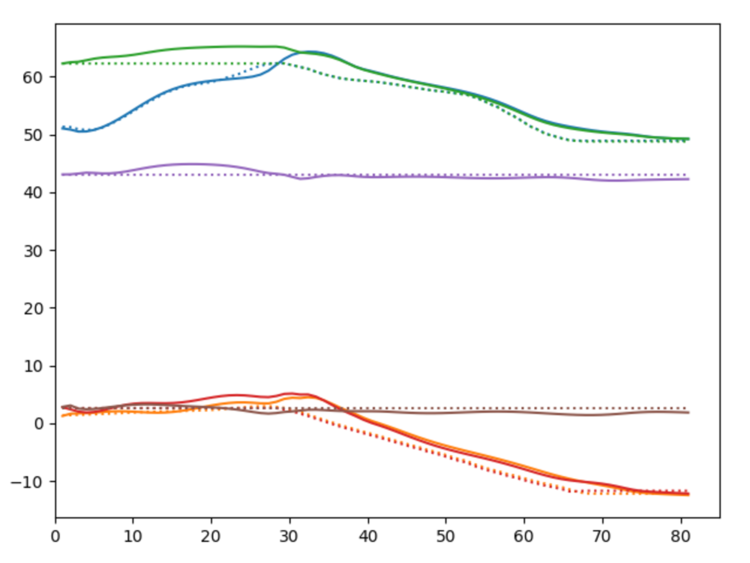}}
\hspace*{\fill}
\subcaptionbox{GLean\label{fig:exp2gpngt}}{\includegraphics[width=0.3\textwidth]{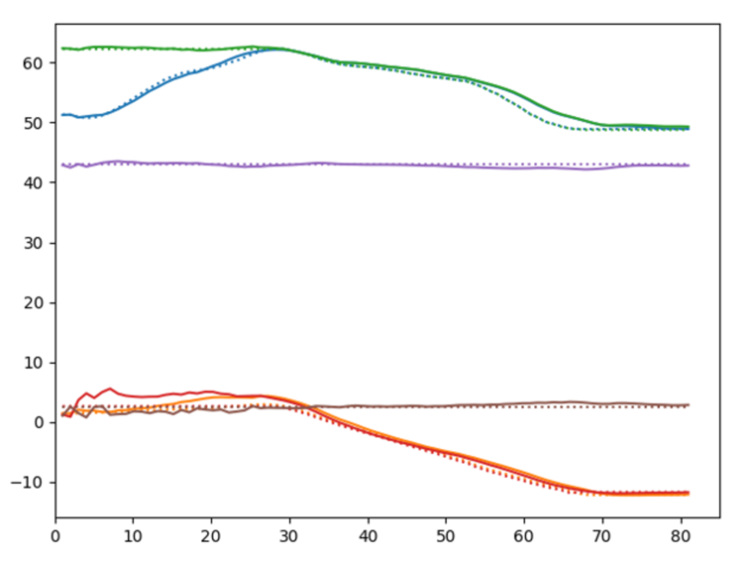}}
\hspace*{\fill}
\vspace{1em}
\caption{Comparison between the generated sensory predictions (solid lines) and the ground truth sensory states (dashed lines) for \textbf{(a)} forward model, \textbf{(b)} stochastic initial state, and \textbf{(c)} GLean\label{fig:exp2compare}}
\end{figure}

Given that GLean and SI were able to generate plans with successful outcomes while FM had no successful plans, it is apparent that the forward model in this condition is not capable of generating goal-directed plans. This is visible in Figure \ref{fig:exp2compare}, showing a comparison between generated sensory predictions and ground truth sensory states, where unlike GLean and SI, FM is unable to find any motor plan in order to generate a plausible sensory prediction.
Naturally, we may ask whether the observed failure of plan generation by FM is due to insufficient sensory prediction capability. In order to examine this, one-step look ahead prediction capability in the three models were compared. In FM, one-step look ahead prediction was generated at each current timestep by providing the ground truth sensory-motor sequence inputs up to the current timestep. The resultant sensory prediction was compared with the ground truth sensory inputs. For GLean and SI, one-step look ahead prediction was generated analogously. Specifically, with GLean, this was done by using error regression for inferring the latent state at each timestep until the current timestep. With SI, the latent state is inferred at the initial timestep only.

\begin{table}[!htbp]
\caption{Comparison of average errors in sensory predictions generated by GLean, FM, and SI when provided with the ground truth motor states} \label{tbl:exp2compare}
\centering
\begin{tabular}{cc}
\toprule
\textbf{Model}                & \textbf{Average RMSE}\\
\midrule
Forward model (FM)            & 0.0119\\
Stochastic initial state (SI) & 0.0107\\
GLean                         & 0.0086\\
\bottomrule
\end{tabular}
\end{table}

\begin{figure}[!htbp]
\setlength{\fboxsep}{0pt}
\centering
\hspace*{\fill}
\subcaptionbox{FM\label{fig:exp2fwogt}}{\includegraphics[width=0.3\textwidth]{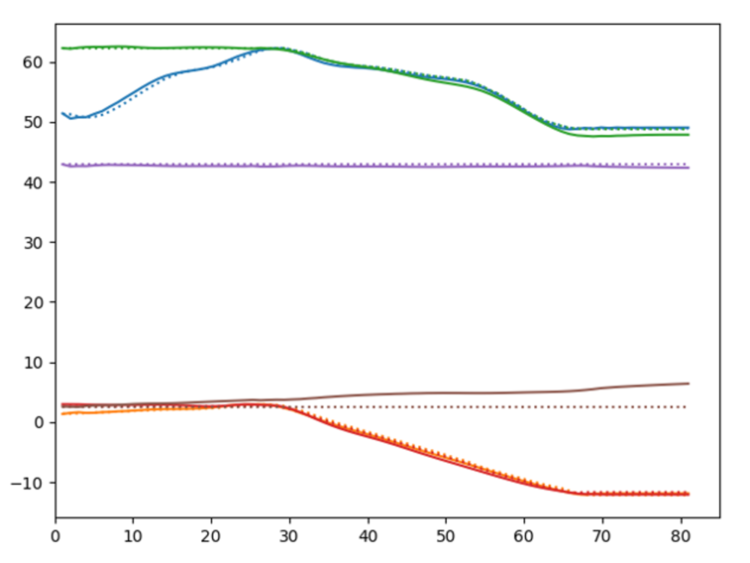}}
\hspace*{\fill}
\subcaptionbox{SI\label{fig:exp2siogt}}{\includegraphics[width=0.3\textwidth]{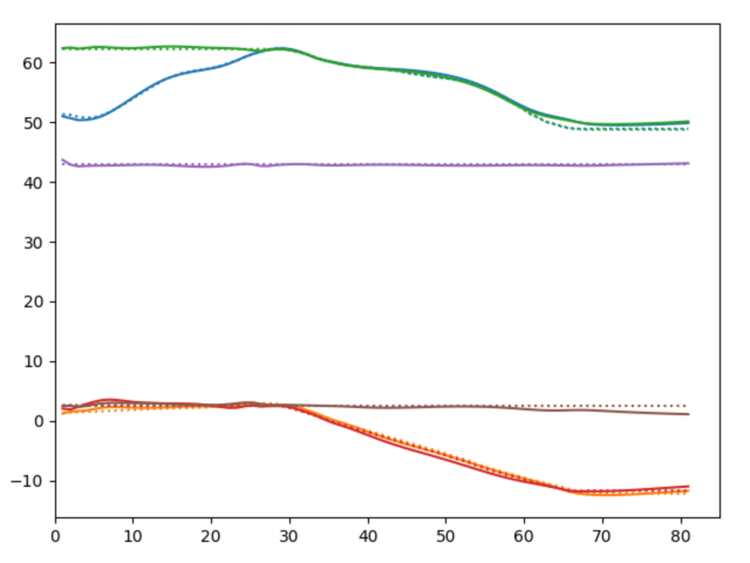}}
\hspace*{\fill}
\subcaptionbox{GLean\label{fig:exp2gpogt}}{\includegraphics[width=0.3\textwidth]{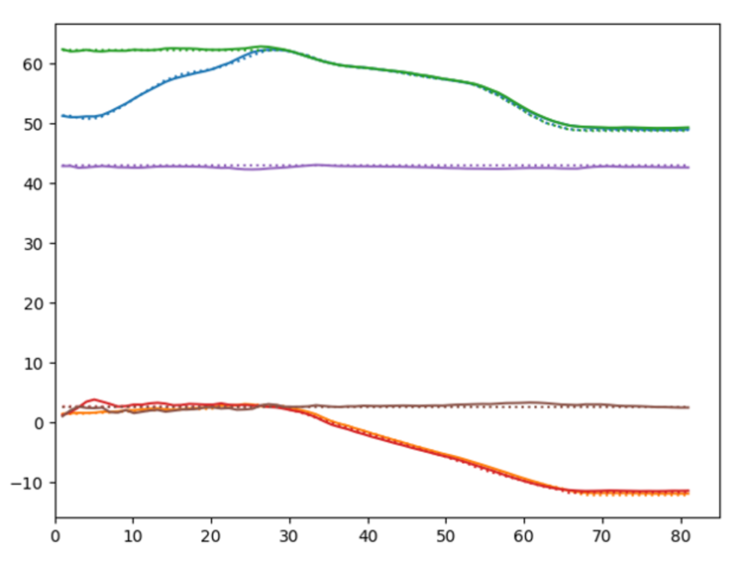}}
\hspace*{\fill}
\vspace{1em}
\caption{Comparison of one-step look ahead sensory prediction (solid lines) and the ground truth (dashed lines) among three different models---\textbf{(a)} forward model, \textbf{(b)} stochastic initial state, and \textbf{(c)} GLean\label{fig:exp2compareo}}
\end{figure}

Table \ref{tbl:exp2compare} shows the result of comparison among those three models. It can be observed that the prediction capabilities of these three models are relatively similar. In particular, by looking at a comparison of one-step ahead prediction for an example trajectory among the models as shown in Figure \ref{fig:exp2compareo}, we observe that FM is able to generate adequate sensory predictions in a similar manner to SI and GLean. This result suggests that the failure of motor plan generation by FM is not due to lack of sensory prediction capability but due to other reasons. We speculate that this is caused by the fact that no prior knowledge or constraints exist for generating motor sequences in FM. We discuss this issue in Section \ref{sec:discussion}.

%%%%%%%%%%%%%%%%%%%%%%%%%%%%%%%%%%%%%%%%%%
\section{Conclusion and Discussion} \label{sec:discussion}
The current study proposed GLean as a novel goal-directed planning scheme to investigate the problem of how agents can generate effective goal-directed plans based on learning using limited amount and range of sensory-motor experiences. GLean was developed using the frameworks of predictive coding (PC) and active inference (AIF). With these frameworks, learning is conducted by inferring optimal latent variables and synaptic weights for maximizing the evidence lower bound. In a similar manner, GLean accomplishes goal-directed planning by inferring optimal latent variables for maximizing the estimated lower bound. Actual implementation of GLean was achieved by using the predictive coding inspired variational recurrent neural network (PV-RNN) previously proposed by our group \cite{ahmadi19}.

The model was evaluated using a simple virtual agent in 2D environment and a more complex simulated robotic pick and place task. The analysis based on the results from the first experiment revealed that the prior distribution developed initial state sensitive deterministic dynamics by increasing the complexity with a strong meta-prior. Meanwhile, it developed noisy stochastic dynamics by reducing the complexity with a weaker meta-prior.

Both experiments showed that GLean produces the best performance in goal-directed planning by achieving sufficient generalization in learning when setting the meta-prior to an adequate intermediate value between the two extremes of weak and strong during the learning phase. Furthermore, it was shown that motor plans cannot be generated for those goals set in unlearned regions. This is because the learned prior tends to prevent the motor trajectory from going beyond the learned region.

The performance of GLean in goal-directed planning was compared against the forward model (FM) and stochastic initial state model (SI) using the robotic pick and place task. The results showed that GLean outperforms the other two models, especially when considering the simulation results. Moreover, it was shown that FM cannot generate corresponding motor plans at all even with sufficient capability for predicting next timestep sensory inputs when provided with the current motor commands. This outcome can be accounted for by the fact that in FM the motor plan search is carried out without any learned priors for constraining generation of hypothetical motor sequences, since FM does not facilitate any functions for probabilistically predicting next motor commands. 

In this circumstance, improbable trajectories that seemingly reach given goals can be generated by arbitrarily combining motor states in sequences that happen to minimize the distal goal error. On the other hand, in the case of GLean a generative model is learned as a probabilistic mapping from the latent state in the current timestep to the proprioception in terms of the joint angles as well as the exteroception in terms of sensory state. In this case, motor sequence plans can be inferred under the constraints of the learned prior by which goal-directed plans can be generated within the boundary of well-habituated trajectories.

The aforementioned idea aligns well with the concept of ``niche construction'' of agents discussed in \cite{kirchhoff18}. It is argued that agents should not attempt to learn complete global models of the world, and instead should learn local models of habituated patterns which should be feasible given that the amount of possible experiences in the world is certainly limited. The free energy minimization principle \cite{friston06} naturally realizes this as its inherent drive for minimizing surprise places limits on plan generation as well as actions beyond the boundary of habituated space. 

Another similar line of research that we have recently become aware of is model-based reinforcement learning for plan generation \cite{ha18, hafner19}. An agent learns either deterministic \cite{ha18} or stochastic \cite{hafner19} latent dynamics for predicting the sensation and reward in a similar manner to FM. The agent after learning can generate actions not by using an action policy network as is the case in model-free reinforcement learning but by planning in the latent space using an evolutionary search for maximizing the reward accumulation in the future.

The aforementioned studies, however, could suffer from the problem of generating faulty action plans because the planning process cannot be constrained by a learned action prior for preventing action space search from going beyond the learned region, as the current paper has discussed. This problem could be solved if the model-based learning and the model-free learning components could be sufficiently combined as the latter could provide the action prior to be former.

The current study has potential to be extended in various directions in future study. One significant drawback in the current study is that an optimal value for meta-prior which results in the best generalization in learning and planning can be obtained only though trial and error. Although our preliminary study showed that generalization is less sensitive to the setting of the meta-prior when an ample amount of training data is used, the meta-prior should still be set within a reasonable range. Future study should explore possible measures for adapting meta-prior automatically depending the training data.

One interesting possibility is that shifts of the meta-prior during planning could affect the quality of motor plan generation analogously to the choking effect \cite{beilock07, cappuccio19}. The choking effect is the tendency of athletic experts to show performance disruption such as drops in the quality and precision of generated sensorimotor behavior. Cappuccio et al. \cite{cappuccio19} proposed that this effect can be accounted for by imprecise precision modulation during active inference for generating motor plans. This imprecise precision modulation could take place during the inference of the latent variable, if the meta-prior in our model is set with different values during the plan generation phase compared to the optimal values used during the learning phase. Future study should explore such mechanisms in detail.

Another drawback is that the current investigation is limited to an offline plan generation processes. Extended studies should investigate how the model could deal with the problem of online planning, which requires the model to be responsive to dynamically changing environments in real time. For this purpose, the model should be extended such that all three processes of (1) recognizing the current situation by maximizing the evidence lower bound, (2) updating current goal-directed plans based on currently recognized situation by maximizing the estimated lower bound, and (3) acting on the environment by executing the plan, to be carried out in real time, as has been demonstrated by a simulation study on the retrospective and prospective inference scheme (REPRISE) \cite{butz19}.

In such a situation, an interesting problem to be considered is how to dynamically allocate cognitive computational resources required for real time computation of these multiple cognitive processes by adapting to the on-going situation under a resource bounded condition. It is also important to investigate how agents can assure the minimum cognitive and behavioral competency for their survival when the optimization involved with these cognitive processes cannot be guaranteed under real time constraints. These research problems are left for future studies.

Although the current study employed supervised training schemes for acquiring the generative models, it may also be interesting if self-exploration-based learning can be introduced to the scheme. One possible scenario for achieving this would be to incorporate the idea of intrinsic motivation \cite{oudeyer07, forestier17} into the model. With intrinsic motivation, agents tend to explore particular goals more frequently for which the success rate of achievement improves more rapidly or other goals \cite{forestier17}. The exploration can switch to other goals when the exploration of the current goal hits a plateau in its improvement. In order to incorporate such a mechanism into GLean, GLean should be extended to facilitate a mechanism for learning a meta-policy \cite{forestier17} to generate its own goals, some more frequently than others, by monitoring the improvement rate for successful achievement of each goal. It may be worthwhile to examine what sort of development in generating various goal-directed behaviors, from simple to more complex, can be observed by using this scheme.

%%%%%%%%%%%%%%%%%%%%%%%%%%%%%%%%%%%%%%%%%%

\section*{Acknowledgments}
The authors are grateful for the help and support provided by Dr. Ahmadreza Ahmadi, as well as the Scientific Computing section of Research Support Division at OIST.

%%%%%%%%%%%%%%%%%%%%%%%%%%%%%%%%%%%%%%%%%%

\bibliographystyle{abbrvnat}
\bibliography{GP.bib}

\end{document}